# KuBERT: Central Kurdish BERT Model and Its Application for Sentiment Analysis

Kozhin muhealddin Awlla
Kma190h @cs.soran.edu.iq
Computer Science Department, Faculty of Science, Soran University, Soran, Erbil, Kurdistan, Iraq

Hadi Veisi
h.veisi@ut.ac.ir
College of Interdisciplinary Science and Technologies, School of Intelligent Systems, University of Tehran, Iran

Abdulhady Abas Abdullah
abdulhady.abas@ukh.edu.krd
Artificial Intelligence and Innovation Centre, University of Kurdistan Hewler, Erbil, Iraq

## Abstract

This paper enhances the study of sentiment analysis for the Central Kurdish language by integrating the Bidirectional Encoder Representations from Transformers (BERT) into Natural Language Processing techniques. Kurdish is a low-resourced language, having a high level of linguistic diversity with minimal computational resources, making sentiment analysis somewhat challenging. Earlier, this was done using a traditional word embedding model, such as Word2Vec, but with the emergence of new language models, specifically BERT, there is hope for improvements. The better word embedding capabilities of BERT lend to this study, aiding in the capturing of the nuanced semantic pool and the contextual intricacies of the language under study, the Kurdish language, thus setting a new benchmark for sentiment analysis in low-resource languages. The steps include collecting and normalizing a large corpus of Kurdish texts, pretraining BERT with a special tokenizer for Kurdish, and developing different models for sentiment analysis including Bidirectional Long Short-Term Memory (BiLSTM), Multi-Layer Perceptron (MLP), and finetuning the BERT classifier. The proposed approach consists of 3 classes: positive, negative, and neutral sentiment analysis using a sentiment embedding of BERT in four different configurations. The accuracy of the best-performing classifier, BiLSTM, is 74.09%. For the BERT with an MLP classifier model, the maximum accuracy achieved is 73.96%, while the fine-tuned BERT model tops the others with 75.37% accuracy. Additionally, the fine-tuned BERT model demonstrates a vast improvement when focused on two 2-class sentiment analyses positive and negative with an accuracy of 86.31%. The study makes a comprehensive comparison, highlighting BERT's superiority over the traditional ones based on accuracy and semantic understanding. It is motivated because several results are obtained that the proposed BERT-based models outperform Word2Vec models conventionally used here by a remarkable accuracy gain in most sentiment analysis tasks.

# 1. Introduction

Sentiment analysis (SA) is rapidly expanding as a research field within natural language processing (NLP) and text classification. In the realm of sentiment analysis, particularly within the context of the Kurdish language, a notable advancement has been achieved through the utilization of Word2Vec (Muhealddin and Veisi, 2022) for analyzing sentiments in the past. Sentiment analysis holds substantial practicality for various entities such as businesses, researchers, governments, politicians, and organizations, offering insights into public sentiments crucial for decision-making processes. To improve the accuracy and detail of sentiment analysis, the proposed method in this study adopts the BERT (Bidirectional Encoder Representations from Transformers) language model for word embedding. This integration will improve the understanding of semantic nuances and contextual intricacies in the Kurdish language. Kurdish is an Indo-European language in the Indo-Iranian branch that comprises North Kurdish (Kurmanji) and Central Kurdish (Sorani) as its main dialects (Abdullah and Veisi, 2022). Kurmanji is the central Kurdish speech community encompassing southeastern Turkey and parts of northern and northeastern Syria, while Sorani is spoken in north of Iraq and western Iran (Abdullah et al., 2024). Although Kurdish has millions of native speakers, resource challenges face it; there are few dedicated tools for processing published texts in the language. In this study, we focus on Central Kurdish; henceforth, Kurdish refers to the Central Kurdish Branch (CKB) (Muhamad et al., 2024). Another general technique involved in sentiment analysis is word embedding, which refers to the construction of vector space presentations of words and documents. The relevance of word embedding comes from the capability to capture both syntactic and semantic relationships among words. An example of word embeddings learning-based methods is the work of Mikolov et al. in 2013 with the Word2Vec and GloVe by Pennington, Socher, and Manning in 2014. These have been taken on by researchers in performing sentiment analysis in their studies, as well as in others (Araque et al., 2017; Tang et al., 2014). In word embedding, BERT greatly enhances due to its consideration of bidirectional context that enables the capturing of a richer understanding of the relationship between words. Unlike most traditional methods like Word2Vec and Global Vectors, BERT learns the representation of a word considering the entire context of the sentence and the words before and after the sentence. The major strength of BERT is its learning ability to understand complicated word relationships with semantics. Due to large-scale pre-training and bidirectional input, BERT can capture subtle nuances and contextual variations in word usage, making it particularly effective for tasks like sentiment analysis (Acikalin et al., 2020).

The primary contribution of this study is the development of a robust text corpus and fine-tuned BERT models for sentiment analysis in Kurdish, addressing a significant gap in NLP resources for this low-resource language. By aggregating over 300 million tokens from diverse sources, this effort not only creates a vital dataset but also advances sentiment analysis capabilities specifically for Kurdish. This substantial advancement in NLP technology significantly enhances the understanding and processing of the Kurdish language, which has been previously neglected in sentiment analysis research.

This paper follows a structured organization outlined as follows: Section 2 provides an overview of related works in the field of sentiment analysis technology. Section 3 focuses on the development and design of sentiment analysis specifically customized for the CKB language. In Section 4, the paper elucidates the proposed method and the architecture of the applied model for the CKB language sentiment analysis system. Section 5 is dedicated to presenting the results and discussions. Finally, Section 6 concludes the paper.

# 2. Related Works

With the growth of online platforms like review sites, blogs, and social networks, the field of sentiment analysis has gained both opportunities and challenges. This development allows individuals to access a wide range of opinions through technology. However, addressing these aspects in the context of Kurdish, a low-resource language, presents unique challenges in terms of data availability, linguistic nuances, and the development of effective sentiment analysis tools. In the following, some related work in the sentiment analysis of the Kurdish language is presented. In the meantime, the section also highlights some of the prior techniques that proved effective in treating low-resource languages in these domains.

## 2.1 Exploring Sentiment Analysis: Methods and Innovations

Sentiment analysis, or opinion mining, is one of the applications of NLP approaches to evaluate the varying opinions and feelings in people (Yang et al., 2024). More fundamentally, and with the perspective shifting to deep learning, models such as the ones described above have dealt with the categorization of emotions through insights. Other approaches include extracting lexical sentiments from documents using Bhowmik et al (2021) and associating features with sentiment using bi-gram and trigram (Tiffani., 2020). In the last few years, more influence has been gathered by deep learning in NLP. Deep learning tasks in NLP, the majority of which are represented by word vector representations, are observed by authors associated with Araque et al. (2017). For instance, the model has been the Embeddings from Language Model (ELMo) (Peng et al., 2019) model, which entailed a generative recurrent neural network such as RNN-LSTM-based unsupervised learning bidirectional to develop contextualized word vectors. Other types of learning algorithms include Word2Vec (Church., 2017) and GloVe, which are helpful in the treatment of text in text classification, clustering, and information retrieval. These word embedding methods have many benefits compared to the old-style bag-of-words representation; for instance, similar words are located close to each other in the embedding space and have smaller dimensionality, which is well explained in (Worth, P.J., 2023). Since 2018, the time factor of NLP has changed rapidly. This is mainly because of the introduction of BERT, proposed in (Devlin et al., 2018). The outstanding effectiveness of BERT in various NLP tasks has led to the incorporation of such a "pre-training plus fine-tuning" approach within the NLP community. This innovation dramatically changed the usual way of dealing with pre-trained word vectors during subsequent NLP tasks. Unlike the conventional methods with word-level vectors as outputs, BERT aims to yield sentence-level vectors, leading to a richer contextual understanding. This kind of sentence-level vector training proves to be much more effective for most NLP tasks. The actual extensive development in language representation that broke free of the LSTM methods was the BERT model. BERT uses attention-based Vaswani et al. transformers, also without recurrence, and this is significant progress in language representation.

## 2.2 Understanding BERT and Its Role in Enhancing Sentiment Analysis

The BERT model represents a major advancement in NLP, revolutionizing how we approach various linguistic tasks, including sentiment analysis. At the core of BERT is the Transformer architecture, which leverages a self-attention mechanism to process and understand text more effectively than traditional models and its general architecture is shown in Figure 1. Unlike earlier language models, which processed text in a unidirectional manner (either left-to-right or right-to-left), BERT is unique in its bidirectional approach. This allows the model to simultaneously consider both the left and right context of each word in a sentence, thereby capturing the complex relationships between words more accurately (Wang et al., 2024).

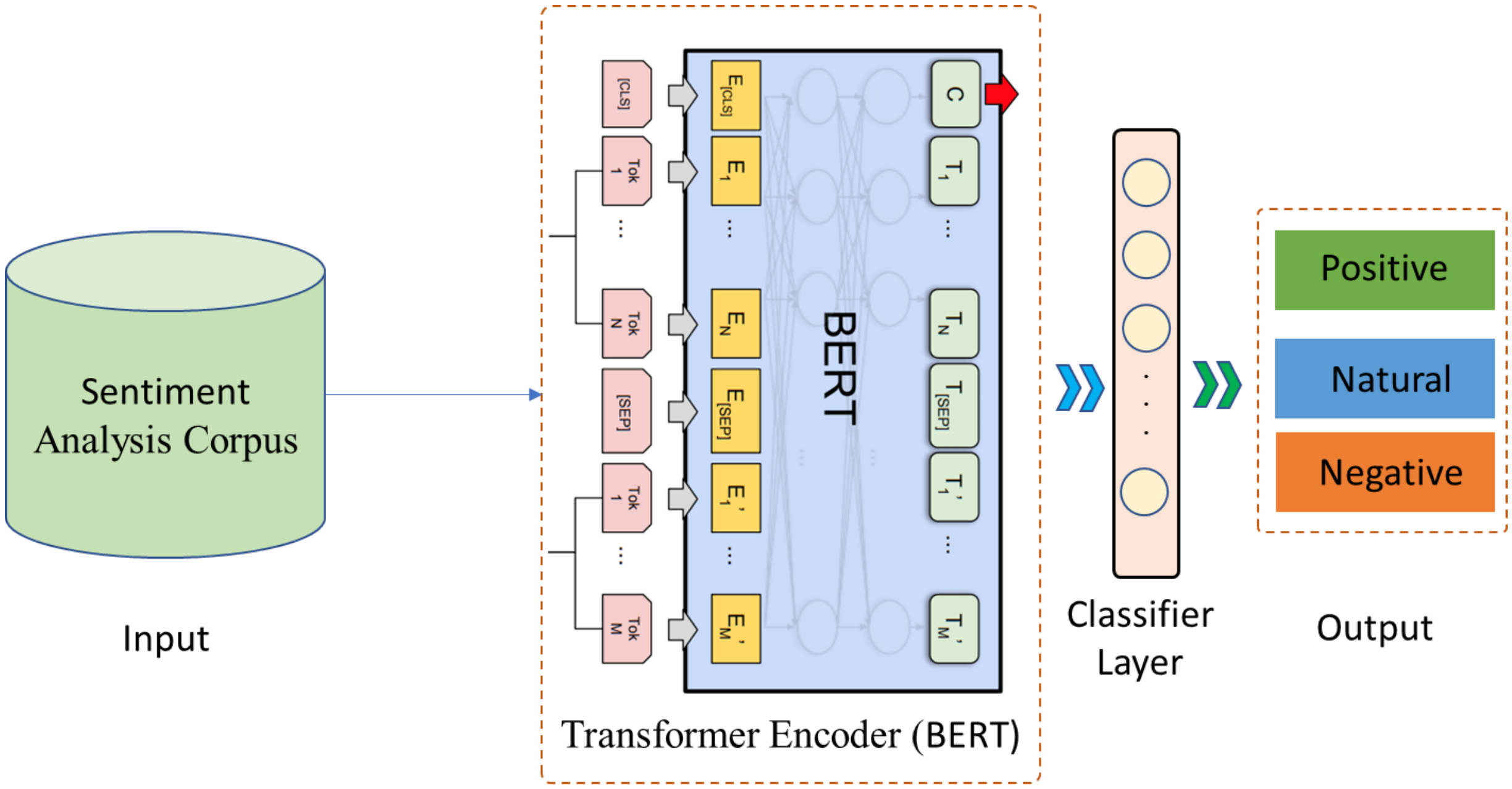

Figure 1 Overall architecture Sentiment Analysis of BERT model

BERT's bidirectional nature is particularly beneficial for sentiment analysis, as sentiment is often conveyed through subtle relationships between words. For example, in a sentence like "I don't think the movie was that bad," understanding the true sentiment requires considering the entire sentence, including how negations and modifiers work together. Traditional models like Word2Vec or bag-of-words approaches often fail to grasp such nuances, as they lack the ability to capture long-range dependencies and contextual meanings. In contrast, BERT excels at these tasks due to its ability to understand both the immediate and global context of words (Tiwari and Nagpal., 2022).

The foundation of BERT is the Transformer model, which relies on the self-attention mechanism. This mechanism enables BERT to weigh the importance of each word in relation to others in the input text, regardless of their distance. Words that are far apart in a sentence can still influence each other's meaning. For sentiment analysis, this is crucial as it allows the model to understand complex sentiments that may be spread across multiple parts of a sentence or paragraph. BERT's ability to autonomously learn these intricate dependencies makes it highly effective in understanding and analyzing the sentiment of various text forms. BERT's support for sentiment analysis extends further through fine-tuning, where a pre-trained BERT model is adapted to specific tasks using labeled datasets. For sentiment analysis tasks, fine-tuning typically involves connecting the BERT output to a classifier, such as a softmax layer, to categorize the text into sentiment classes (e.g., positive, negative, or neutral). During fine-tuning, the BERT model adjusts its parameters through backpropagation, optimizing its performance for the specific sentiment analysis dataset (Yu et al., 2017). This adaptability allows BERT to outperform traditional sentiment analysis models by a significant margin, especially when dealing with domain-specific or complex language structures.

In short-text sentiment analysis, BERT proves to be particularly effective. The model can capture the full context of a short sentence and generate a rich, context-aware representation of the entire sentence. This ensures that even in short snippets of text, which are common in social media posts, reviews, and comments, BERT can accurately identify emotional nuances. Whether it's identifying sarcasm, negations, or complex emotional expressions, BERT's deep contextual understanding enhances the overall performance of sentiment analysis tasks. Moreover, BERT's flexibility across multiple languages, combined with its ability to fine-tune on various types of data, makes it suitable for sentiment analysis across diverse domains and languages. Whether analyzing product reviews, social media sentiment, or movie ratings, BERT consistently achieves higher accuracy by leveraging its ability to model intricate language relationships. In domains where sentiment is heavily influenced by context, such as customer feedback or opinion mining, BERT's bidirectional attention mechanism ensures a more accurate and nuanced understanding of the sentiment expressed (Mamta and Ekbal., 2024).

In conclusion, BERT's integration of the Transformer architecture, with its self-attention mechanism and bidirectional context analysis, has transformed sentiment analysis by providing a more precise and contextually aware understanding of language. This powerful model not only improves performance on short-text sentiment tasks but also extends its capability to complex and domain-specific sentiment analysis applications.

## 2.3 Sentiment Analysis in the Age of BERT

Sentiment analysis is a task of classification that entails determining the polarity of sentiment. In the field of emotion analysis, emotion representation models are typically divided into two main categories: categorical and dimensional emotion analysis. Categorical sentiment analysis involves representing sentiments or opinions using discrete, discrete categories such as positive, negative, or neutral. This approach simplifies emotions into predefined layers, facilitating tasks such as opinion extraction or emotion classification. On the other hand, dimensional sentiment analysis goes beyond discrete categories and uses continuous scales to measure emotions across multiple dimensions. These dimensions typically include valence, which reflects the positivity or negativity of an emotion, and arousal, which refers to the intensity or level of activation of the emotion. By considering a broader range of emotional nuances, dimensional models provide a more accurate understanding of emotions. Introducing both models in the related work section will help frame the development and application of sentiment analysis approaches in different areas of research and practice (Bordoloi and Biswas., 2023). Many approaches, from classical to state-of-the-art deep learning techniques, have been applied to deal with this challenge. With this motivation in mind, we present below an overview of the work in the field of sentiment analysis, emphasizing the work in which BERT was applied. It is articulated by Ansar et al. (2021) in this manner: "The proposed rules in this paper to extract the aspect-opinion phrases have been able to reduce the sentence length by 84% and complexity by 50%." They proposed a redefined version of TF-IDF to identify important aspects, along with a novel technique for word representation that captures both local and global contexts. They use the pre-trained BERT for sentiment classification to overcome its classical sequence length limitations and increase efficiency and accuracy by 8%. The effectiveness of their work is shown with a comprehensive analysis of nationwide movie reviews and Sainthood.

A result of the inadequacy of work for Turkish sentiment analysis research and the insufficiency of Turkish NLP resources, their work presents two new approaches. End. They employed the first approach for fine-tuning a pre-trained multilingual BERT model. They indicated that the second BERT fine-tuning primary approach, on Turkish text translated to and from English, significantly improved over known methods. They evaluated these approaches on the Turkish movie and hotel review datasets and created a labeling for each review, either as positive or negative. The results show that these approaches based upon BERT give very high accuracies, with the highest using the movie review dataset and outperforming the current methods for this task.

For instance, Jafarian et al. (2021) push forward ABSA with the Persian Pars-ABSA dataset. It emphasizes how a pre-trained BERT model is effective, pointing out the advantages of sentence-pair input in ABSA. Results show that fusing the pre-trained Pars-BERT (Pouromid et al., 2021) model with a Natural Language Inference auxiliary sentence (NLI-M) significantly boosts ABSA accuracy. This method attains an accuracy of up to 91%, with a significant absolute gain of 5.5% over prior state-of-the-art results on the Pars-ABSA dataset.

The research work demonstrated by Islam et al. applies to sentiment analysis in the Bengali language, which is characterized by extensive inflectional variations. This adds new manually tagged 2-class and 3-class SA datasets for Bengali and uses a multi-lingual BERT model with fine-tuning by transfer learning for sentiment classification. This attains a rate of 71% in the 2-class classification, an improvement over the present 68%, and sets the benchmark for the first 3-class classifier at 60%.

A detection system is proposed in this study that uses a deep convolutional neural network along with BERT to identify and classify offensive posts of both monolingual and multilingual formats in social media. They investigate several approaches to multi-linguality that include collaborative and translation-based methods. In this study, with the use of different BERT pre-trained word-embedding techniques against both Bengali and English datasets, a better performance of 91.83% was obtained as compared to existing offensive text classification algorithms. Sentiment analysis is critical to get an insight into public opinion, especially with the exponential increase in online data. This paper introduces the Double Path Transformer Network (DPTN) for holistic review categorization; it fuses global and local information based on a parallel design of self-attention and convolutional networks. By sharing knowledge optimization (GSK), the model attains 95% accuracy, displaying excellent management of class imbalances in large textual datasets (Kumar et al., 2024).

## 2.4 Comparative Sentiment Analysis for Related Languages

To fully appreciate the implications of BERT on sentiment analysis for Central Kurdish, it is essential to examine its applications in other linguistically similar languages, such as Persian and Arabic. These languages share structural and contextual similarities and many of its phonemes and alphabet with Kurdish, making them ideal benchmarks for comparison in sense analysis tasks. In recent research on sentiment analysis for Persian, different models such as Support Vector Machine (SVM), Naive Bayes and MLP have been widely used, although their independent performance often falls short in obtaining high-quality results. To address these limitations, the work proposes a hybrid model combining Naive Bayes, a rule-based system and BERT, specifically targeting sentiment analysis on Persian Twitter data. The hybrid approach outperforms the traditional models, achieving an accuracy of 89%, exceeding the standalone BERT accuracy of 86% (Vakili et al., 2024). This study deals with the extraction of moral sentiment from Persian text using moral foundation theory, presenting a manually annotated data set of 8000 Persian tweets. The study evaluates several models, finding that the refinement of the pre-trained Persian BERT model outperforms other models, with a proposed energy conversion-based model giving competitive results while providing faster inference (Karami et al., 2023). another paper analyzes Persian text sentiment using the BERT algorithm and compares its performance with past approaches. The results show that ParsBERT significantly outperforms other models, achieving an F1 score of 96.62 and an accuracy of 94.38 on the original dataset (Zardak et al., 2023).

This paper analyzes Arabic tweets related to the visit of the Chinese President to Saudi Arabia and the convening of related summits with Arab nations, using CAMeLBERT's sentiment analysis model. It indicates that most of the tweets are neutral, following the media line, with an excess of positive ones over negative ones, which mirrors positive Arab public perception. Positive tweets bring out cooperative and partnership potentials. In contrast, negative tweets are on the significant strategy concerns Americans and Iranians, subservience to China, and overtime-on-overtake of the interests of the Chinese over others. These are very keen insights that will help solve some of the challenges in the future relations between China and the Arabs (Inoue et al., 2021). Another study introduces FusionAraSA, a fusion majority voting technique combining four models (AraBERTv1, MARBERT, Modified MARBERT, and AraBERTv2) for Arabic sentiment analysis. Achieving over 95% accuracy and F1 score on two datasets, FusionAraSA demonstrates exceptional performance and highlights the potential of model fusion for Arabic sentiment analysis tasks (Alawi et al., 2024). This study addresses aspect-based sentiment analysis (ABSA) by jointly identifying aspect terms and their sentiments using a multi-task learning approach. The model integrates BERT, BiGRU, attention, and a CRF layer, achieving superior performance on an Arabic hotel dataset compared to baseline models (Bensoltane and Zaki., 2025).

## 2.5 Central Kurdish Sentiment Analysis

This will discuss the state of sentiment analysis for the Kurdish language at present, along with modern language processing technologies. At present, though much has been initiated to enrich the Kurdish language, it remains in the first stage of its development. Due to this fact, there is a massive gap between the present stage of Kurdish sentiment analysis and the capacities of contemporary language processing technologies. Research Gap and Bridging the Gap in Developing Kurdish Sentiment Analysis The paragraph highlights the existing gap, and at the same time, the field in question needs further development and research to exploit sentiment analysis in the Kurdish language fully. Abdulla and Hama. 2015 pioneered sentiment analysis for CKB language using a naive Bayes classifier. They employed a bag-of-words technique, where they processed 15 000 text documents, 8 000 of them being positive and the rest negative. The corpus was obtained from social networks like Facebook, Twitter, and Google+. They had a performance of the approach that led to an accuracy measure of 0.72, indicating the various learning methods of Kurdish sentiment analysis. This study focuses on Kurdish hate speech detection, using a dataset of 6882 Facebook comments labeled into hate and non-hate categories. SVM, Decision Tree (DT), and Naive Bayes (NB) algorithms were applied and compared. The results show that SVM achieved the best performance, with an F1 score of 0.687, marking an important step in hate speech detection for the Kurdish language (Saeed et al., 2022). Amin et al. In 2022, they showed the challenges of adapting the techniques of sentiment analysis to the Kurdish language. The problems involved in these challenges range from collecting data and taking out relevant features to making classifications. Overcoming these challenges, the authors brought two different approaches: one based on machine learning and the other based on lexicon-based techniques. In these studies, authors analyze to bridge the gap in sentiment analysis between English and the Kurdish language on social media. The research first curates and annotates a new dataset for sentiment analysis of Kurdish. It then evaluates a range of machine learning algorithms, such as the Naïve Bayes philosophy-based, against deep learning techniques like ANN, LSTM, and CNN. However, the most exciting discovery is that the performance of the Naïve Bayes outperformed all other models, with an accuracy of 78%. This article delves into sentiment analysis to detect sentiment on specific topics of COVID-19 and online education in the Kurdish language regarding social platforms of the Kurdistan Region of Iraq. Furthermore, in 2022, Muhealddin and Veisi also participated in the line of Kurdish sentiment analysis with 14,881 comments stemming from various Facebook pages for data creation. Based on this data, they achieved an accuracy of 71.35% with an LSTM classifier using Word2vec embeddings. In another paper by Daneshfar in (2024), the authors describe the making and labeling of a dataset for sentiment analysis in Central Kurdish. The authors explore approaches that employ classical machine learning algorithms and neural networks for sentiment analysis and further leverage transfer learning using pre-trained models to enhance the dataset in this task. Their research enlightened the fact of performing data augmentation to gain fantastic F1 scores or accuracy, although the task is inherently complex. The KurdiSent dataset is an annotated dataset with over 12,000 instances used for the sentiment analysis of the Kurdish language, which is annotated to two classes: positive and negative. As stated in the experiment results, the XLM-R model obtains a good value of 85% for precision, outperforming the methods based on machine learning and deep learning classifiers (Badawi et al., 2024).

Table 1: This study compared with earlier studies on sentiment analysis for Central Kurdish.

| Years | Dataset | Techniques | Performance | Reference |
|---|---|---|---|---|
| 2015 | 15,000 Facebook, tweets and google | Naive Bayes classifier | 66% accuracy | Abdulla et al., 2015 |
| 2022 | 20000 websites, facebook and tweeter | SVM and Naïve Bayes | - | Amin et al., 2022 |
| 2022 | 6882 Facebook comments | SVM, DT and NB | 68.7% F1 | Saeed et al., 2022 |
| 2022 | 18,000 Facebook comments | Word2Vec, LSTM | 71% accuracy | Muhealddin and Veisi., 2022 |
| 2023 | 6,408 tweets | LSTM and CNN | 78% accuracy | Mahmud et al., 2023 |
| 2023 | 5685 tweets | BiLSTM | 61% accuracy | Daneshfar., 2024 |
| 2024 | KurdiSent | Transformers (BERT) | 85% accuracy | Badawi et al., 2024 |

Table 1 Overview of past studies, datasets used, and methodology applied for sentiment analysis of the Kurdish text. While these have provided many valuable insights, researchers have mainly been using the available methods without harnessing the powerful and novel capabilities of recent advances in transformers for developing more sophisticated sentiment analysis tools for the Kurdish language. As a result, in the presented study, we aim to conduct a benchmark analysis for advanced sentiment classification in the Kurdish language using recently developed language models, such as BERT. It will be essential to increase the NLP capabilities in the Kurdish language, as LMs based on Transformers have state-of-the-art performance in text understanding and generation. We are, therefore, seeking that the implementation of the use of LMs be used for Kurdish sentiment analysis to allow the state of the art in this field to grow and allow more precision and subtleties in sentiment analysis applications associated with this language.

# 3. Proposed Method: Central Kurdish BERT for Sentiment Analysis

In this section, we discuss how the sentiment analysis for the central Kurdish language is performed. After describing the text corpus, we explain our usage of BERT for word and document representation in Central Kurdish, called KuBERT. Unlike most other language models, where tokenization of the input text must be pre-trained, BERT requires its tokenizer, namely the WordPiece Tokenizer of Song et al. This was followed by the training of four different BERT models, which helped develop sentiment analysis models with three classifiers BiLSTM, MLP, and fine-tuning. As a result, general Structure of the proposed method shown in figure 2.

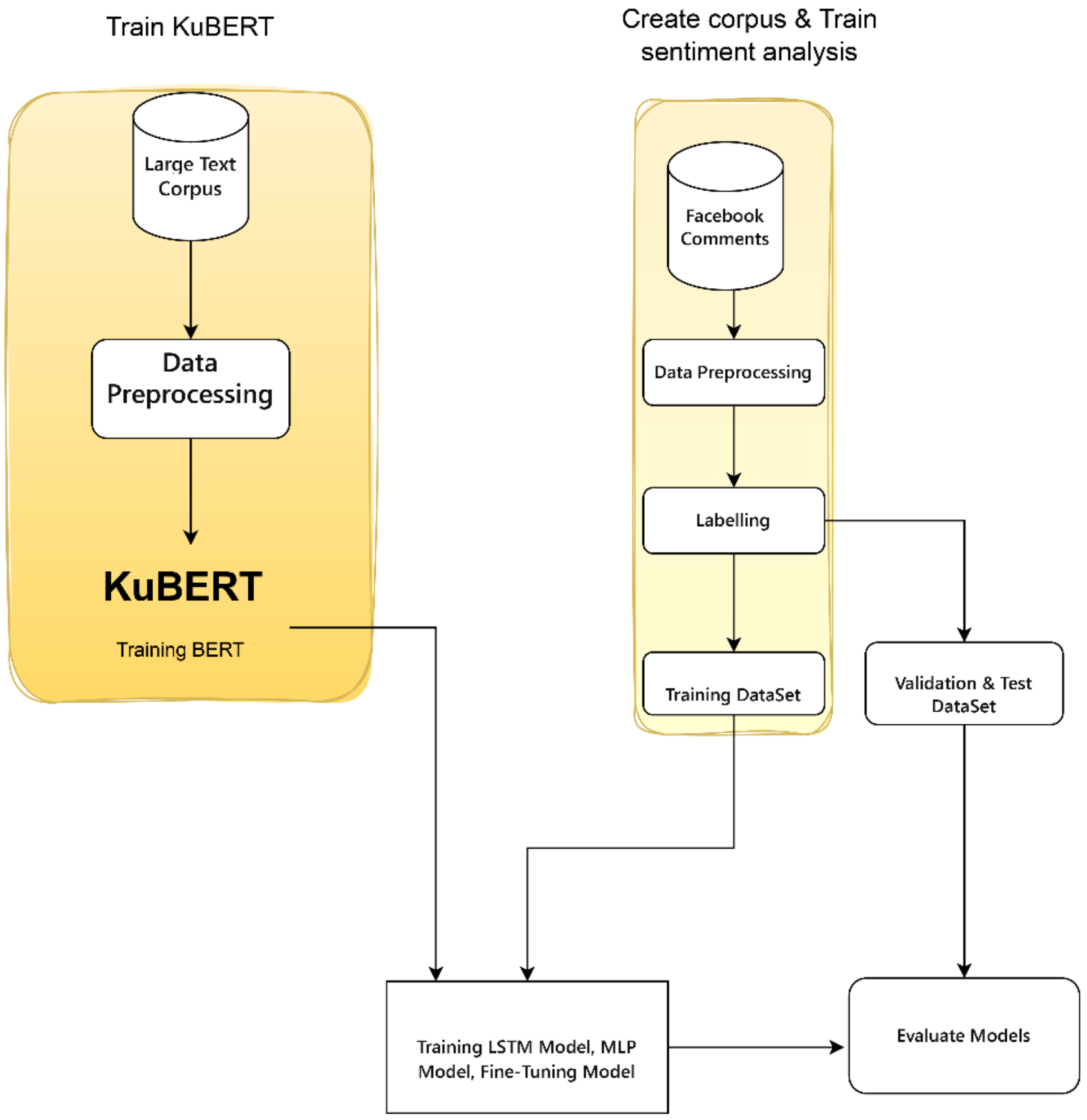

Figure 2 Structure of the proposed sentiment analysis for the Central Kurdish using BERT

## 3.1 Dataset

### 3.1.1 Text Corpus for BERT Language Model

Acquiring large amounts of data is the primary challenge in training deep learning models, and it can be particularly difficult for languages with low processing power. This is since large datasets are required for these sophisticated models to train well. As a result, there aren't many models available for under-resourced languages like Kurdish, mostly because it's hard to get enough data. Due to the dearth of digital tools, collecting data in Kurdish is far more difficult than in most other languages. To build word vectors and train a BERT for Kurdish, information was gathered from three primary sources. The AsoSoft corpus (Veisi et al., 2020) had more than 188 million words that were collected from a range of sources, such as websites, textbooks, and magazines (Table 2). AramRafeq and Muhammad Azizi also contributed by gathering information from Kurdish websites, totaling over 60 million tokens (Table 3). With 48.5 million words, the Oscar 1 2019 corpus was the last one to be used.

Table 2: AsoSoft Kurdish Text Corpus (Veisi, et al., 2020)

| Source | Number of tokens |
|---|---|
| **Crawled From Websites** | **95M** |
| **Textbooks** | **45M** |
| **Magazines** | **48M** |
| **Sum** | **188M** |

Table 3: Muhammad Azizi and Aram Rafeq Text Corpus[2]

| Source | Number of tokens |
|---|---|
| **Wikipedia** | 13.5M |
| **Wishe Website** | 11M |
| **Speemedia Website** | 6.5M |
| **Kurdiu Website** | 19M |
| **Dengiamerika Website** | 2M |
| **Chawg Website** | 8M |
| **Sum** | 60M |

Therefore, we have used these three corpora, a total of 296.5 million tokens to train the word embedding network shown in Table 4.

Table 4: The Kurdish text corpus used in this paper to train the BERT language model

| Corpus name | Number of tokens |
|---|---|
| **Oscar 2019 corpus** | 48.5M |
| **AsoSoft corpus** | 188M |
| **Muhammad Azizi and AramRafeq corpus** | 60M |
| **Sum** | 296.5M |

### 3.1.2 Dataset for Sentiment Analysis

Muhealddin and Veisi (2022) documented the data collection process in detail. Initial data was collected from Facebook, an online social media platform, by selecting posts designed to capture user attention.

[1] https://huggingface.co/datasets/oscar-corpus/oscar
[2] https://github.com/DevelopersTree/KurdishResources/

From this, a total of 13 popular pages were chosen, yielding 18,450 comments for analysis. Following data collection, a comprehensive cleaning and normalization process was undertaken to remove noisy data and ensure a focus on the core Kurdish language. Each comment was then classified into one of three emotions (positive, negative, or neutral) by three independent annotators. Given the rich rhetorical nature of the Kurdish language, this labeling process was executed with great care. As summarized in Table 5, the final refined dataset consists of 14,881 high-quality comments covering a wide range of topics.

Table 5 The number of samples of the collected dataset

| **Topics** | **No. of data** |
|---|---|
| Sports | 509 |
| Economy | 1,112 |
| Entertainment | 2,056 |
| Education | 713 |
| Technology | 1,252 |
| Lifestyle | 2,015 |
| Fashion | 807 |
| Food | 1,265 |
| Travel | 100 |
| Religious | 865 |
| journalist | 1,714 |
| Art | 511 |
| Political | 1,962 |
| **Sum** | **14,881** |

As shown in Table 6, the average sentence length of our dataset is 11 tokens, and the total number of tokens is more than 151K.

Table 6 Data statistics of the Central Kurdish SA dataset

| **Topics** | **Value** |
|---|---|
| **Longest Sentences** | 512 tokens |
| **Average Sentences Length** | 11 tokens |
| **Total Words** | 151,889 |

The summarized results of this dataset are presented in Figure3.

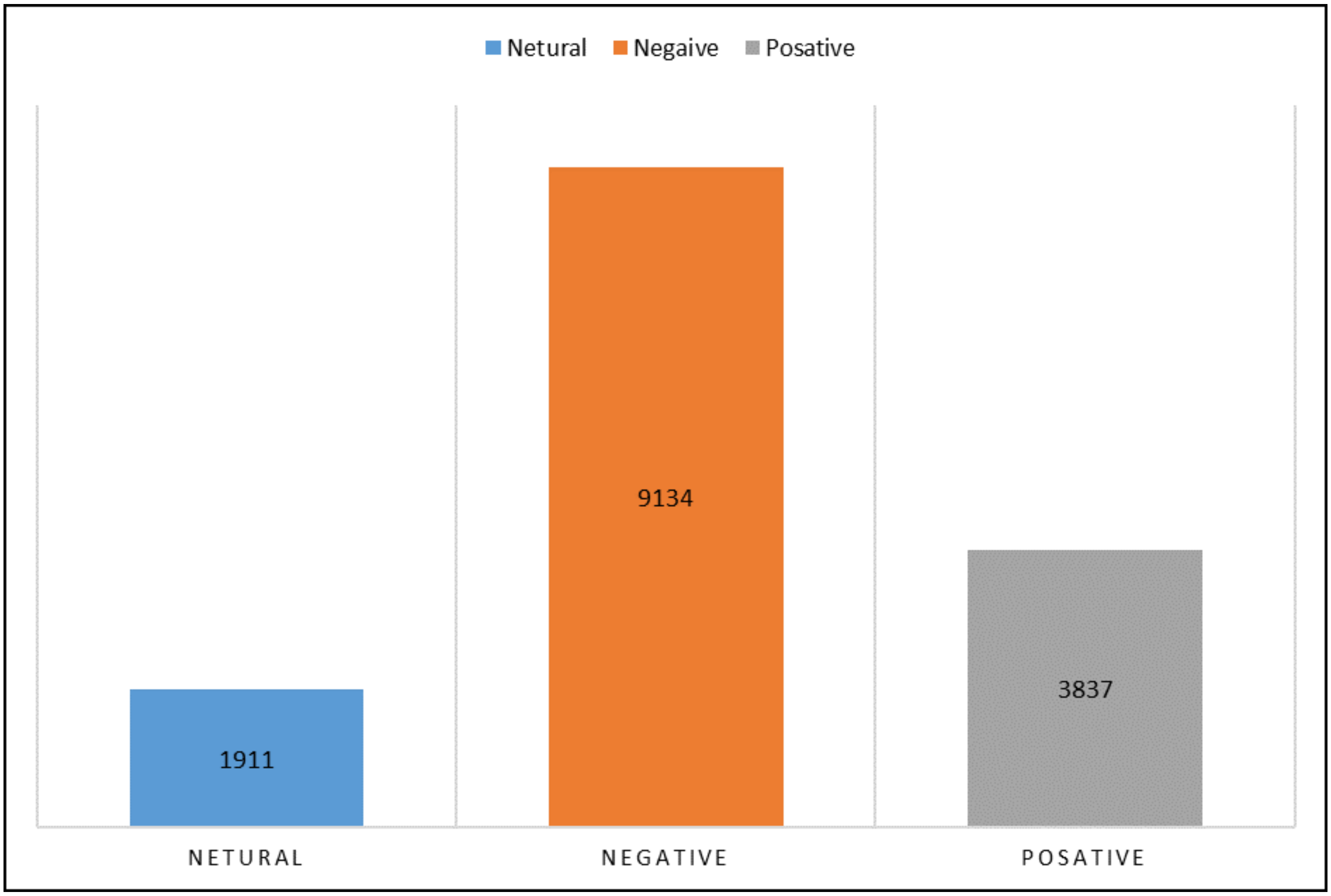

Figure 3: Distribution of the 3-class SA corpus for the Central Kurdish dataset (Muhealddin and Veisi., 2022)

### 3.1.3 Data Cleaning and Normalizing

The importance of cleaning and normalizing data sets is no less important than collecting the datasets. Therefore, the most crucial step after collecting data is the data pre-processing. Still, this step for the Kurdish language is a hard step to achieve clean and standard data. With an appreciation for AsoSoft[3] (Mahmudi, et al., 2019) being able to create a strong normalizer consisting of a set of applications. The following steps are done in this normalizer.

[3] https://github.com/AsoSoft/AsoSoft-Library-py

- Replace URLs and Emails.
- Replace non-Unicode fonts with Unicode style.
- Separate digits from letters.
- Delete non-Kurdish lines.
- Trim white spaces.
- Remove Empty lines.
- Normalize characters
- Remove the non-Kurdish text
- Replace initial ر with ڕ
- Normalize punctuations

## 3.2 Central Kurdish Tokenizer

Tokenization is an essential procedure utilized to divide text into more manageable components, commonly known as tokens, which may consist of a single word. This represents a crucial preliminary stride within the domain of NLP. A tokenizer is required for two purposes in this study: to train BERT and to guarantee that sentences intended for sentiment analysis are accurately tokenized before their input use. When sentences are introduced without the use of an experienced tokenizer, a problem called "out-of-vocabulary" arises (Wu et al., 2016), which signifies that the model is confronted with words that are foreign to it. Failure to resolve this issue may result in a substantial degradation of the efficacy of the models. To tackle this issue, sub-word embeddings were generated using the WordPiece tokenizer. This allowed for the encoding of tokens in conjunction with representations, which were then utilized to extract sentiment features from the context.

For instance, phrases like (“نەخێرررررر" means "nooooooo" and "بەڵێێێێێێێێێێ" means "yessssss”) are informal writing words that are exact common in informal text messages by social networks texts, such as Facebook, Twitter, and Youtube. They could be tokenized into sub-word pieces (e.g., بەڵ, ##ێێێێ, ##ێ, and نەخ, ##ێر، ##ررر) where the symbols “##” means the token isn’t at the front of the word. As a result, the model might capture the emotional benefit information from the ("##ێێێێ").

Additionally, in Table 7 given below, the WordPice tokenizer displayed that tested in one sentence and segment sentence in a good case. That tokenizer is trained by nearly 300 million words for 50,000 vocab sizes. This tokenizer is availble on AsoSoft’s github page4.

Table 7: Original sentences (1) and the tokenized sentences using WordPiece sub word tokenization (2)

| | |
|---|---|
| هانی بەکارهێنانی وزەی دۆستی ژینگە بدەن لە چالاکییەکانی ڕۆژانەدا وەک لە میانی چێشت لێنان، ئامرازەکانی گەرم کردنەوە و روناک کردنەوەدا، لە شارەکاندا بەکارهێنانی ئۆتۆمبێلی تایبەت کەمبکرێتەوە و پەرەبدەن بە بەکارهێنانی گواستنەوەی گشتیی | 1 |

[4] https://github.com/AsoSoft/KuBERT-Central-Kurdish-BERT-Model

| | |
|---|---|
| هانی , بەکارهێنانی , وزەی , دۆستی , ژینگە , بدەن , لە , چالاکییەکانی , ڕۆژانەدا , وەک , لە , میانی , چێشت , لێنان , ، ,<br>یامرازەکانی , گەرم , کردنەوە , و , ر , ##ونا , ##ک , کردنەوە , ##دا , ، , لە , شارەکاندا , بەکارهێنانی , یۆتۆمبێلی , تایبەت ,<br>کەمبکرێتەوە , و , پەرە , ##بدەن , بە , بەکارهێنانی , گواستنەوەی , گشتیی | 2 |

## 3.3 KuBERT: Central Kurdish BERT Model

In this study, training BERT model for Central Kurdish using a text corpus, which encompasses 300 million tokens and employs a tokenizer designed to handle 50,000 tokens, is done which requires a strategic approach to parameter configuration. This endeavor involves the development and comparison of four distinct BERT models, each tailored with specific parameter settings as detailed in Tables 6, 7, 8, and 9, respectively. These settings include critical variables such as epochs, iterations, hidden size, vocab size, number of attention heads, number of hidden layers, and batch size, which collectively influence the model's learning efficiency, capacity to understand complex language patterns, and overall performance.

For the first model, referred to as Model1 in Table 6, the parameter values are chosen to provide a baseline understanding of how the Kurdish language can be processed and modeled effectively. This involves setting an appropriate number of epochs and iterations to ensure sufficient exposure to the data without causing overfitting. The hidden size and vocab size are calibrated to capture the linguistic nuances of Kurdish, balancing the need for a comprehensive vocabulary against the limitations of computational resources. The number of attention heads and hidden layers are configured to enable the model to discern intricate language structures and contextual meanings. Lastly, the batch size is selected to optimize the training process, considering the trade-off between computational demand and learning effectiveness.

Subsequent models (1 through 4) specified in Tables 8, experiment with variations in these parameters to explore different aspects of model performance. Adjustments to epochs and iterations might test the model's learning curve over more extended periods or more intensive training cycles. Alterations in hidden size and vocab size could investigate the effects of increased model complexity or a more expansive linguistic range. Variations in the number of attention heads and hidden layers offer insights into how changes in model depth and attention mechanisms impact the understanding of contextual relationships within text. Lastly, experimenting with batch size across models provides data on how different training scales influence model accuracy and efficiency.

This systematic approach to training four BERT models for Kurdish language processing not only aims to optimize each model's performance on the text corpus but also contributes valuable insights into the effective tuning of deep learning parameters for language-specific applications. Through careful adjustment and comparison of these parameters, the project seeks to identify the most effective configurations for accurately modeling the complexities of Kurdish text. The trained BERT[5] models are available on AsoSoft's GitHub page[6].

Table 8: Central Kurdish BERT Parameter Models

| Parameters | Model1 | Model2 | Model3 | Model4 |
|---|---|---|---|---|
| **epochs** | 10 | 20 | 10 | 20 |
| **Itrations** | 1,000,000 | 2,000,000 | 1,000,000 | 2,000,000 |
| **hidden_size** | 384 | 384 | 768 | 768 |
| **vocab_size** | 50,000 | 50,000 | 50,000 | 50,000 |
| **num_attention_heads** | 12 | 12 | 12 | 12 |
| **num_hidden_layers** | 6 | 6 | 6 | 6 |
| **batch_szie** | 12 | 12 | 12 | 12 |
| **GPU** | yes | yes | yes | yes |

## 3.4 Central Kurdish Sentiment Analysis Models

For constructing a sentiment analysis model, we utilize cleanly labeled data. This data is partitioned into two segments: 80% is allocated for the training phase, and the remaining 20% is reserved for testing purposes. Within the scope of the research, the development of the sentiment analysis model employs three prevalent methodologies or classifiers: Fine-Tuning, BiLSTM, and MLP (Multi-Layer Perceptron). For each methodology, we explore the performance of two distinct BERT model variants.

### 3.4.1 Fine-Tuning BERT

In the process of BERT language model training, alongside model generation, Fine-Tuning is implemented using the labeled dataset, as detailed in the previous section. Initially, as illustrated in Figure 2, the BERT model undergoes retraining with the newly labeled dataset and is subsequently evaluated using a test set to assess the model's performance. For every classifier model mentioned, two variants of BERT models are trained, with specifics provided in the ensuing Table 9.

Table 9: Fine-Tuning sentiment analysis model properties for four types of BERT the model

| Parameters | Values |
|---|---|
| Epochs | 3 |
| Max_len | 256 |
| Learning_Rate | 1.00E-05 |
| DropOut_Rate | 0.3 |
| batch_szie | 8 |
| GPU | yes |

[5] https://colab.research.google.com/drive/10iGjJwzt259WR8IXhJHoI_QnKQ7gMW7t#scrollTo=pEsX5SBZSdbx
[6] https://github.com/AsoSoft/KuBERT-Central-Kurdish-BERT-Model & https://huggingface.co/asosoft/KuBERT-Central-Kurdish-BERT-Model

### 3.4.2 BiLSTM Classifier

The BiLSTM serves as one of the classifier options we are considering. The operational schema of this classifier is depicted in Figure 3. Here, the weights derived from the labeled dataset through the BERT model are fed into the BiLSTM classifier alongside their respective labels. This classifier's architecture is composed of three BiLSTM layers followed by a fully connected layer that outputs to three classes, reflecting the number of categories to be classified. For models 1 and 2, we adopt a similar architectural framework but vary the feature size, as illustrated in Figures 4 and 5, respectively. Additional specifications and hyperparameter details for these network configurations are provided in the subsequent Table 10.

Table 10: BiLSTM sentiment analysis model hyperparameters for both types of BERT models

| Parameters | Values models 1 and 2 | Values models 3 and 4 |
|---|---|---|
| Epochs | 3 | 4 |
| Max_len | 256 | 256 |
| Learning_Rate | 1.00E-05 | 1.00E-05 |
| DropOut_Rate | 0.3 | 0.3 |
| batch_szie | 8 | 8 |
| GPU | yes | yes |

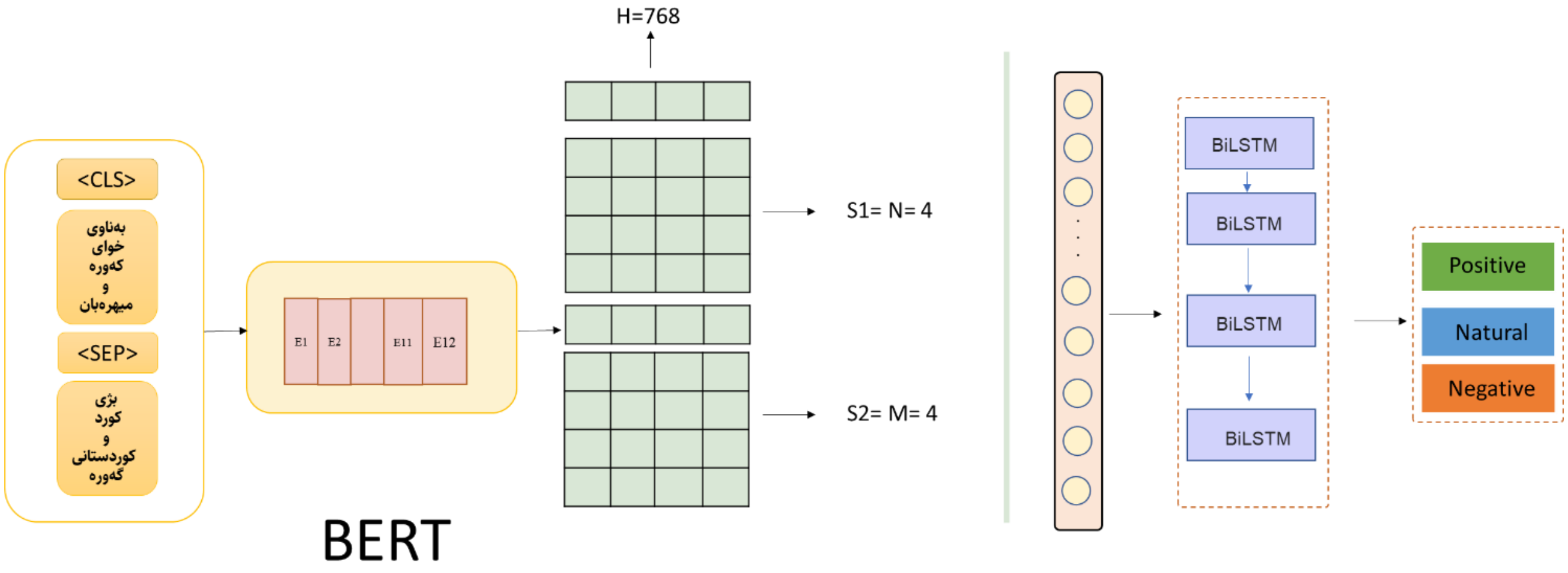

Figure 4: BiLSTM sentiment analysis model for BERT models 3 and 4

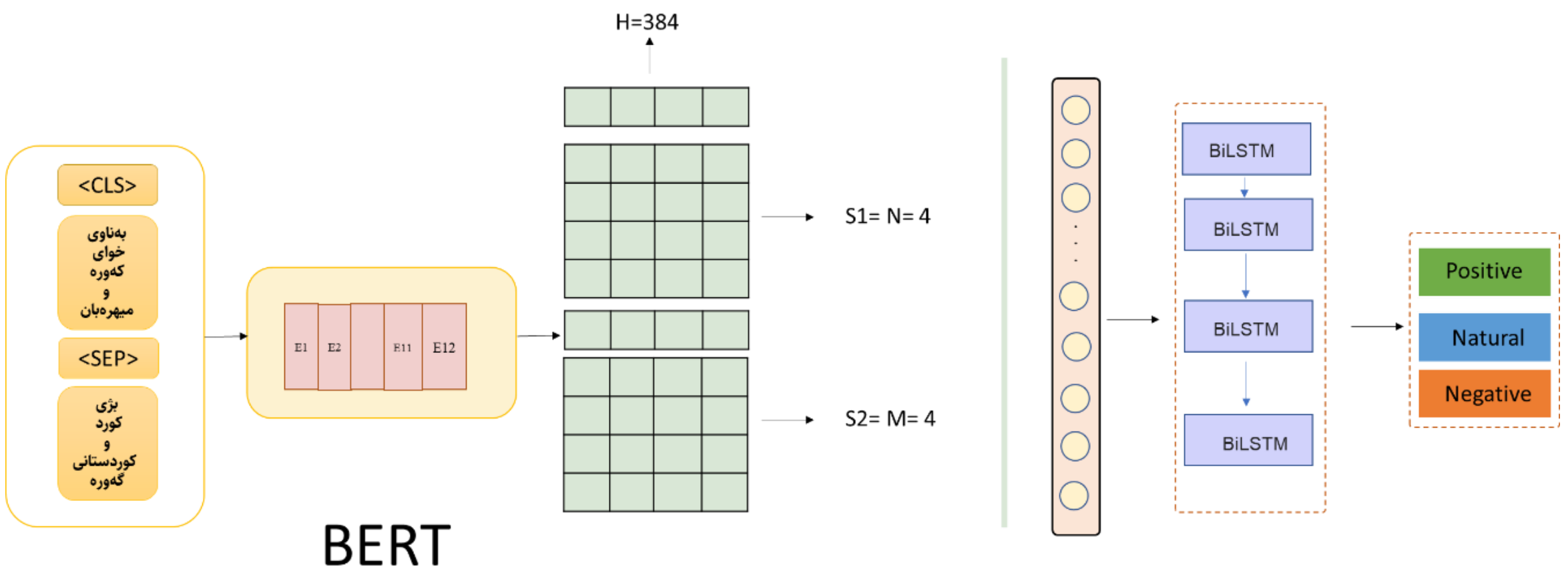

Figure 5: BiLSTM sentiment analysis model for BERT models 1 and 2

### 3.4.3 MLP Classifier

In this work, we proposed to use an MLP classifier for sentiment analysis that makes predictions for positive, natural, and negative classes. We use a small MLP network consisting of 2 hidden layers and an output layer. Relu activation for the hidden layers and SoftMax activation for the final layer are used. Also, 0.3 for dropout rate in the intermediate layer and the Adam optimizer are utilized during backpropagation. The structure of this model is shown in Figure 6. The Table 11 explains more details of the hyperparameters of the networks.

Table 11: MLP sentiment analysis model hyperparameters for BERT models.

| **Parameters** | **Values** |
|---|---|
| Epochs | 4 |
| Max_len | 256 |
| Learning_Rate | 1.00E-05 |
| DropOut_Rate | 0.3 |
| batch_szie | 8 |
| GPU | yes |

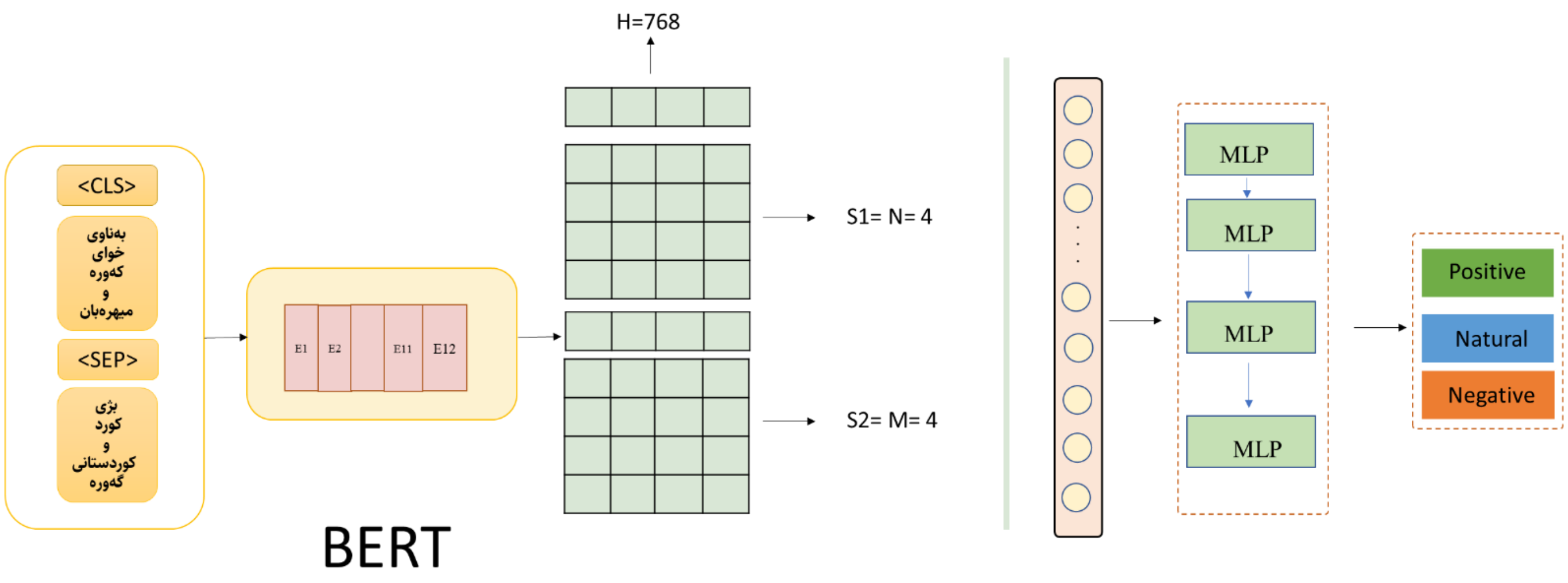

Figure 6: MLP sentiment analysis model for BERT models

## 3.5 Implementation Environment

For the exercises, a Linux machine with an Intel I9 9900K @ 2.10GHz CPU, 500 GB SSD memory, and an RTX 3080 10GB Gainward Ghost GPU with 10 GB memory was used. The implementation environment is Python 3.9, and the packages are torch 1.5, tokenizer 0.7, transformers 2.11, Tensorflow, numpy, pandas, sklearn, gensim, and matplotlib.

## 3.6 Evaluation Metrics

To assess the sentiment analysis task, we used the accuracy and F1 Score evaluation metrics. The result for processing an instance by a sentiment analysis system, can either be:

True Positive (TP) - system prediction is positive, as well as the real value.

False Positive (FP) - system prediction is positive, and the real value is negative.

False Negative (FN) - system prediction is negative, and the real value is positive.

True Negative (TN) - system prediction is negative, as well as the real value.

Accuracy can be computed as the overall correctness of the system i.e., the decisions that the system got right divided by the total number of decisions made by the system. For a binary classification problem:

$$Accuracy = (TP + TN)/(TP + TN + FP + FN) \quad (1)$$

For binary or multi-class classification problems, accuracy remains the most popular evaluation metric in practice. The consistency of the produced solution is evaluated by accuracy based on the percentage of accurate predictions over total instances, as seen in the previous study (Hossin, et al., 2011). The complementary accuracy metric is the error rate that calculates the solution made by its proportion of inaccurate assumptions. In truth, both metrics have been commonly used by academics to discriminate and pick the right solution. The benefits of accuracy or error rate are that, with less difficulty, this metric is easy to compute; applicable to multi-class and multi-label issues; easy-to-use scoring; and easy to understand by humans. However, because our data is completely unbalanced, it is important to use another type of metric called the F-measure. For each of our classes, we find the F-measure to further ascertain the effects of the unbalanced dataset on the models. Also, the micro average is F-measure to each class multiple to weights of each class.

$$Precision\ (p) = \frac{TP}{TP + FP} \tag{2}$$

$$Recall\ (r) = TP/(TP + TN) \tag{3}$$

$$F - Measure = \frac{2 * p * r}{p + r} \tag{4}$$

Where F is f-measure and W is the weight of each class in the sum of all classes.

# 4. Results and Discussion

In this section, we present the results of our evaluations and discuss them in detail. The evaluation was performed using multiple datasets, including the KurdiSent dataset, a well-known benchmark for sentiment analysis in Kurdish. By testing on diverse datasets, we demonstrate the robustness of BERT-based models in addressing sentiment analysis in the core Kurdish language. Furthermore, we compare the performance of different classifiers (Fine-Tuning, BiLSTM, and MLP) and highlight significant advances over traditional models such as Word2Vec.

## 4.1 Results of BERT Word Embedding

### 4.1.1 BERT Model1

The details of this model have been explained in Table 6, the model was trained for more than 2 days on a powerful computer. It has 10 epochs and 1M iteration. As shown in Figure 7, fine-tuning can achieve the best results for the first model of the BERT with an accuracy of 74.29%, followed by MLP with an accuracy of 73.89%, and then BiLSTM with an accuracy of 73.15%. As can be seen from the results, the BERT model fine-tuning had the best results. Figure 8 also shows the F-measures for each of the (MLP, BiLSTM, and Fine-Tuning) models for all classes. Micro average for MLP is 0.72, for BiLSTM is 0.71 and Fine-Tuning is 0.76. As in all models, the neutral class performed low for two main reasons, the first reason is that very little of our data is in the neutral class, that is about 12 percent. The second reason is that neutral text is much more comprehensive than positive and negative, so much more data is needed to train better.

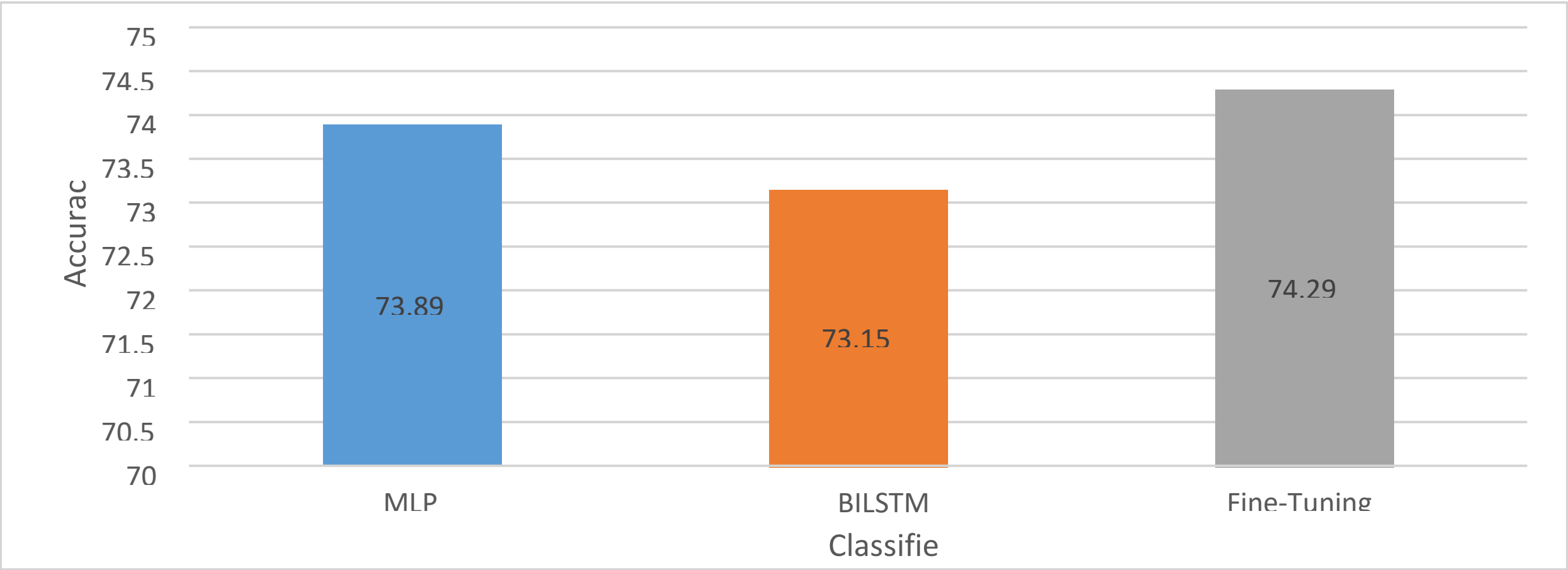

Figure 7: Accuracy of various classifiers for BERT Model1

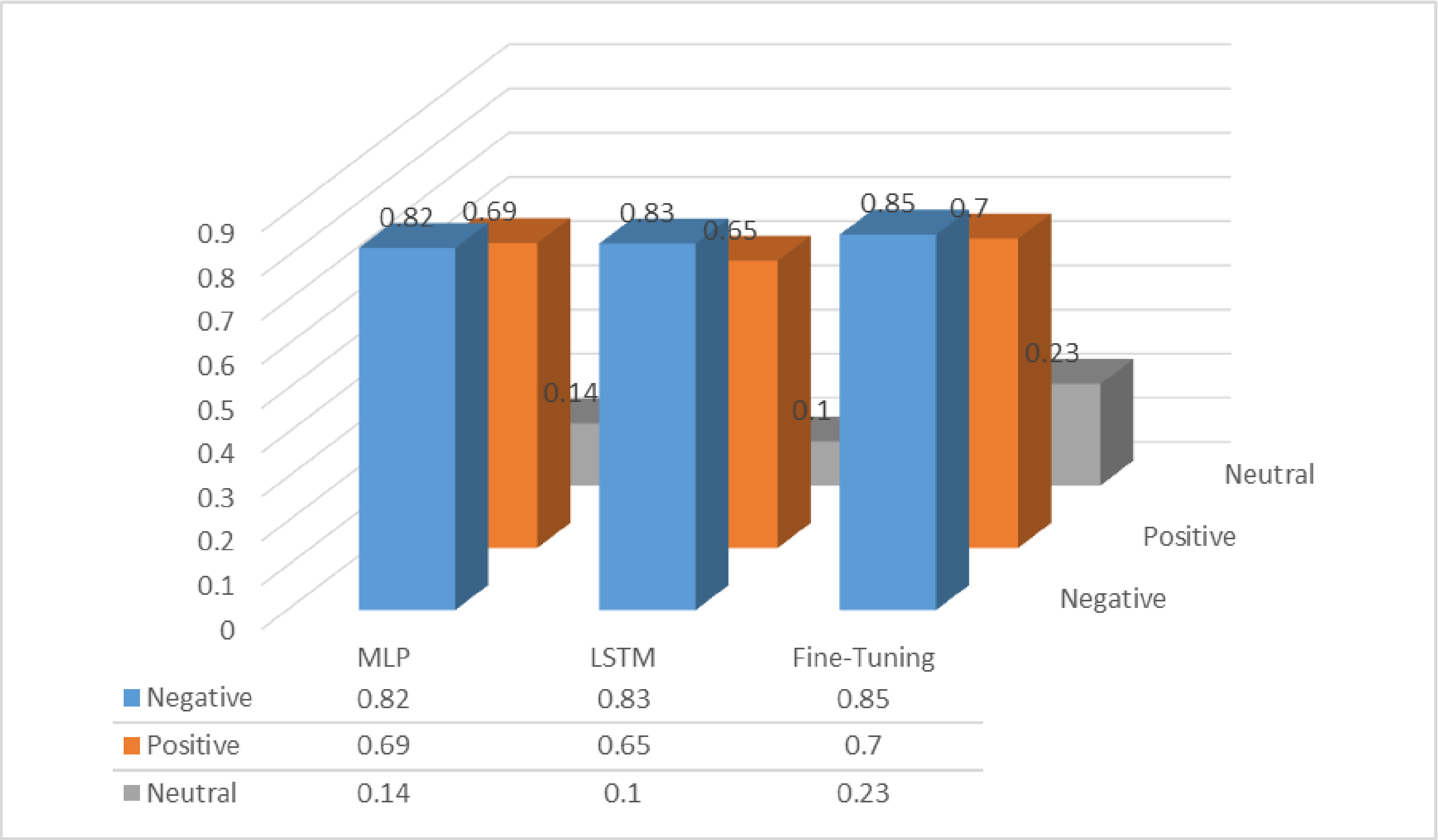

Figure 8: F1 measure of various classifiers for BERT Model1 for each class

### 4.1.2 BERT Model2

The details of this model have been given in Table 7. The training for this model was more than 4 days on the powerful computer. It has 20 epochs and 2M iterations. As shown in Figure 9, fine-tuning can achieve the best results for the first model of the BERT with an accuracy of 74.19%, followed by BiLSTM with an accuracy of 73.69%, and then MLP with an accuracy of 73.12%. As can be seen from the results, the BERT model fine-tuning had the best results, also the BiLSTM classifier improved its performance, but MLP decreased accuracy in this BERT Model2. Figure 10 also shows the F-measures for each of the (MLP, BiLSTM, and Fine-Tuning) models for all classes. The micro average for MLP is 0.72, for BiLSTM is 0.72 and Fine-Tuning is 0.75.

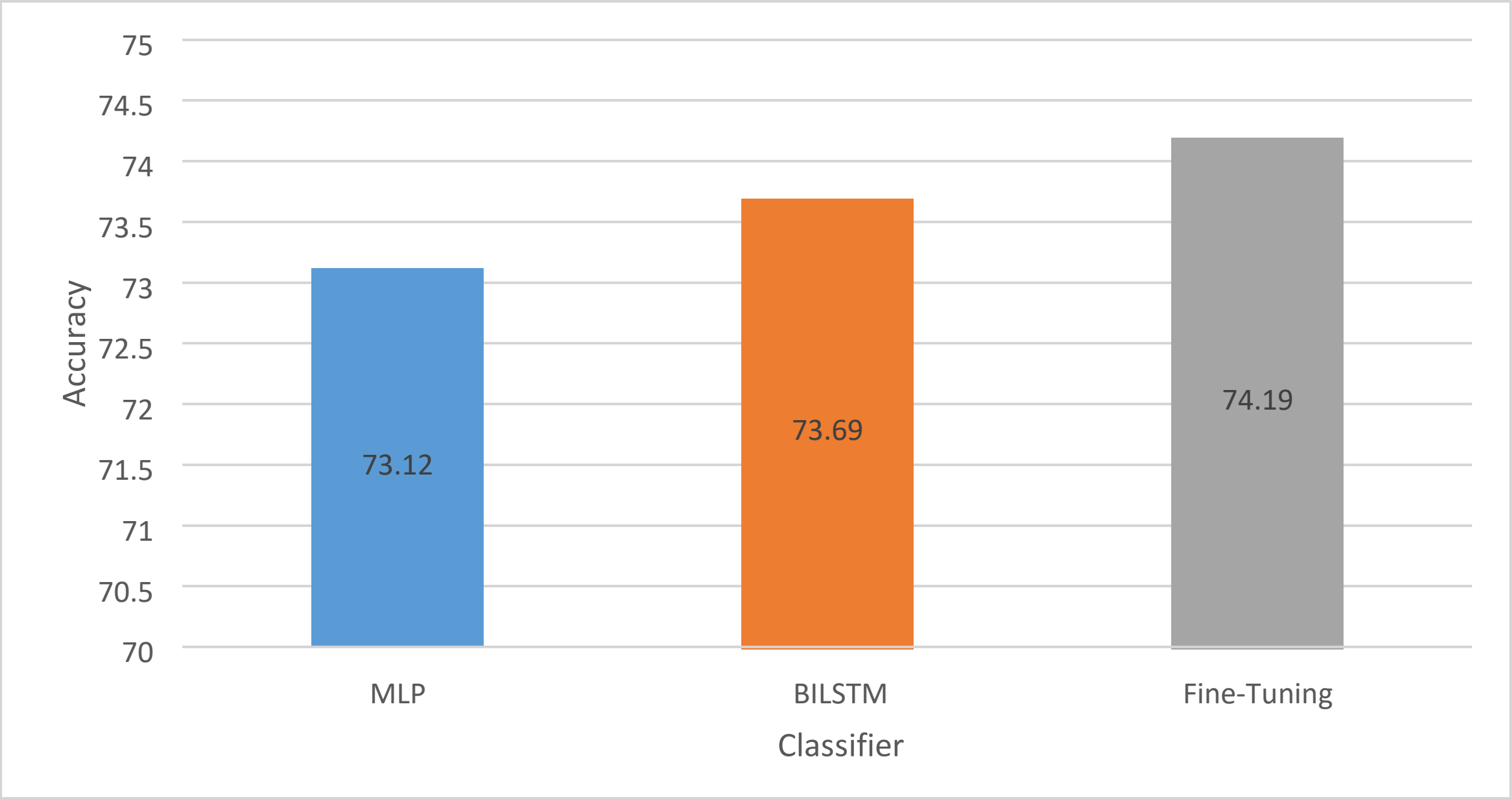

Figure 9: Accuracy of various classifiers for BERT Model2

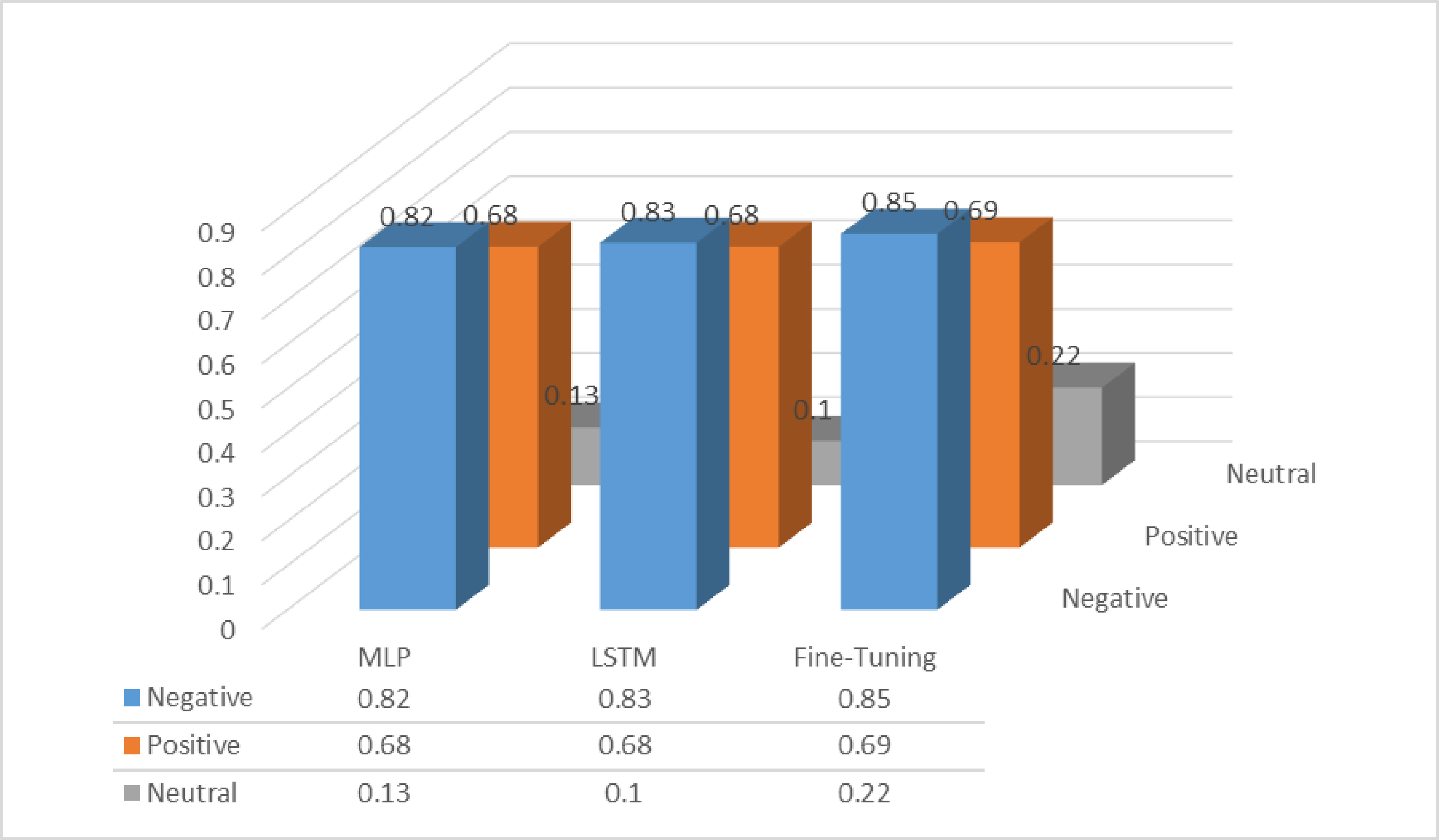

Figure 10: F1 measure of various classifiers for BERT Model2 for each class

### 4.1.3 BERT Model3

The details of this model have been given in Table 8, the model was trained for more than 3 days which has 10 epochs and 1M iterations. As shown in Table 8, the hidden state is interpreted as BERT's normal of 768. As shown in Figure 11, all classifiers showed better performance than in the previous models. As with the previous models, the Fine-tuning showed the best accuracy of 74.9%, the BiLSTM showed a better performance of 74.09%, and the MLP improved on its predecessor with 73.19% accuracy like the first model. Figure 12 also shows the F-measures for each of the (MLP, BiLSTM and Fine-Tuning) models for all classes. The micro average for MLP is 0.71, for BiLSTM is 0.74 and Fine-Tuning is 0.75.

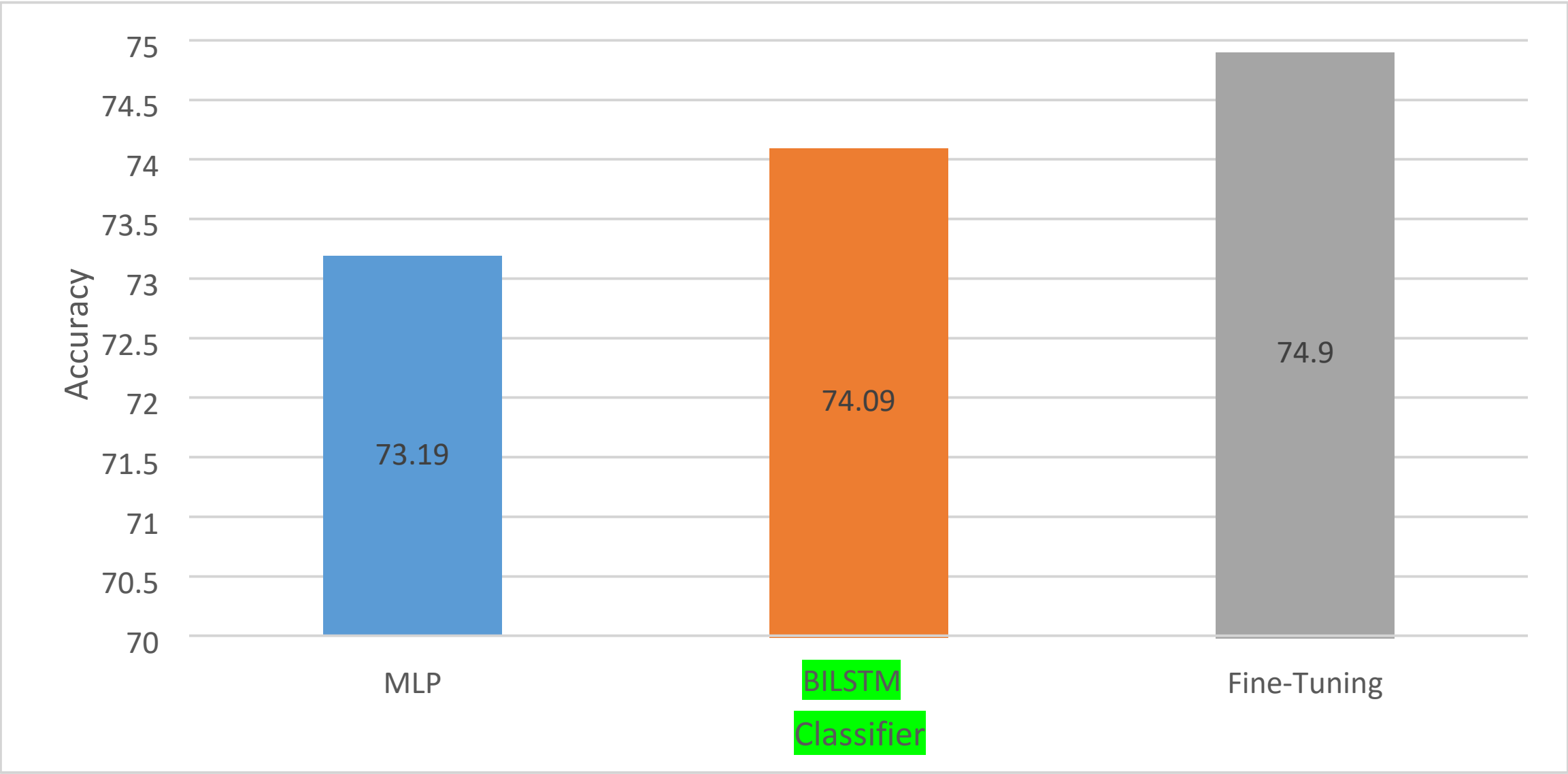

Figure 11: Accuracy of various classifiers for BERT Model3

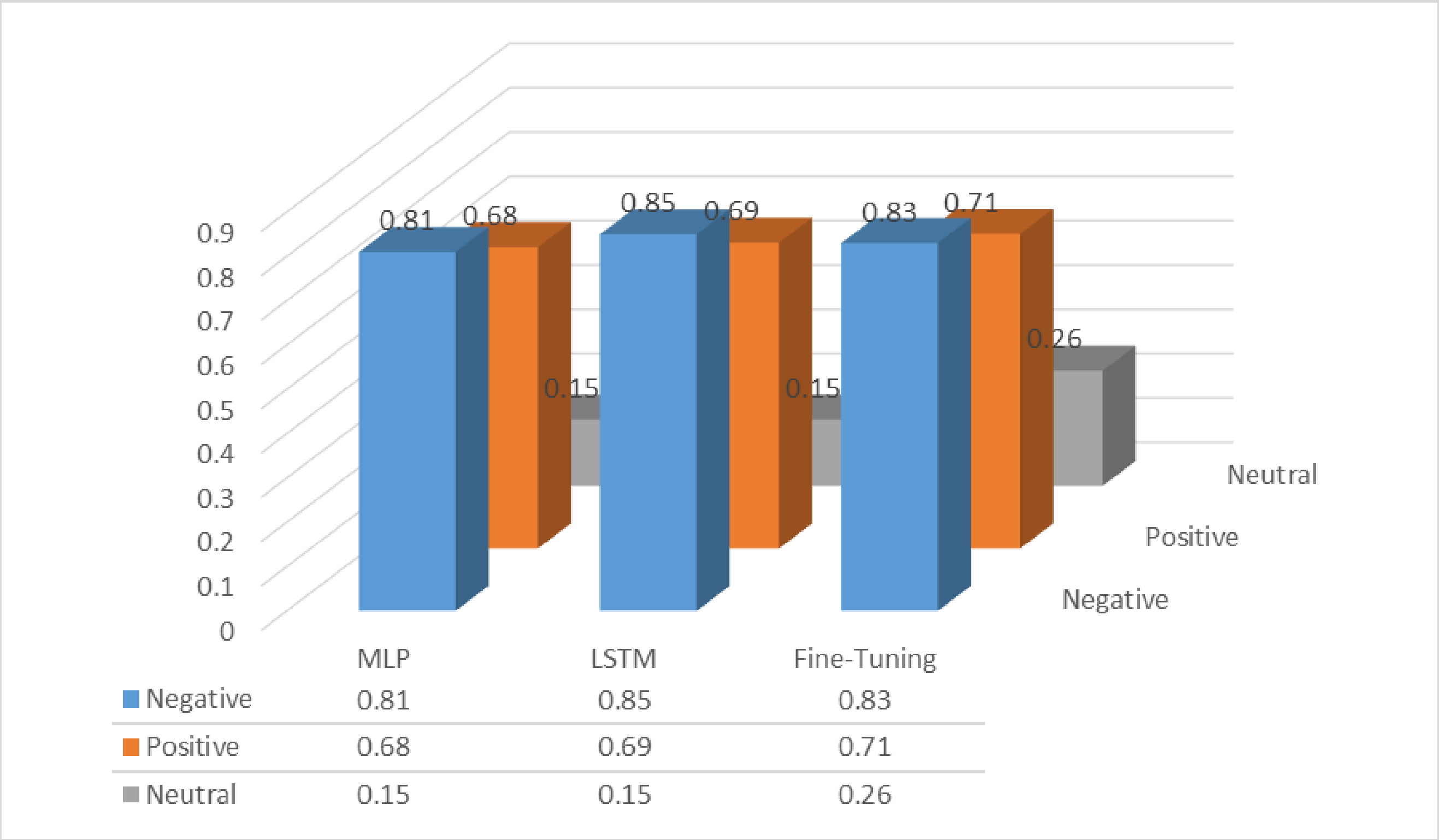

Figure 12: F1 measure of various classifiers for BERT Model3 for each class

### 4.1.4 BERT Model4

In Table 9 this model has been explained. The model was trained for more than 5 days and has 20 epochs and 2M iterations. In this model, BiLSTM decreased its performance while, in addition to that, all other classifiers provided better results. Interestingly, fine-tuning achieved very good results as well as showing better performance than previous results. Fine-tuning accuracy reached 75.37% and MLP was able to show its best case of 73.96%, but BiLSTM accuracy dropped to 73.52% as shown in Figure 13. Figure 14 also shows the F-measures for each of the (MLP, BiLSTM and Fine-Tuning) models for all classes. The micro average for MLP is 0.73, for BiLSTM is 0.73 and Fine-Tuning is 0.78.

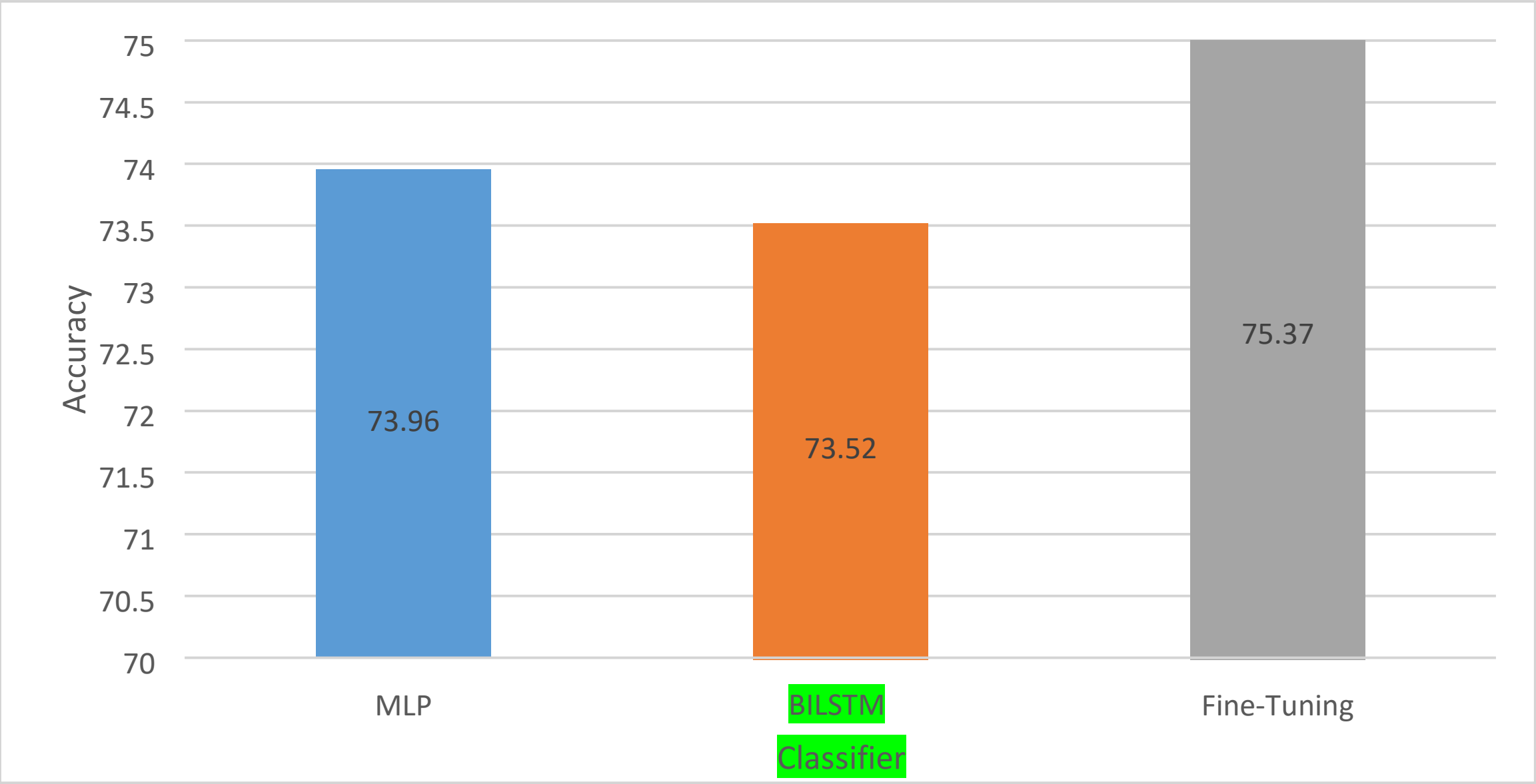

Figure 13: Accuracy of various classifiers for BERT Model4

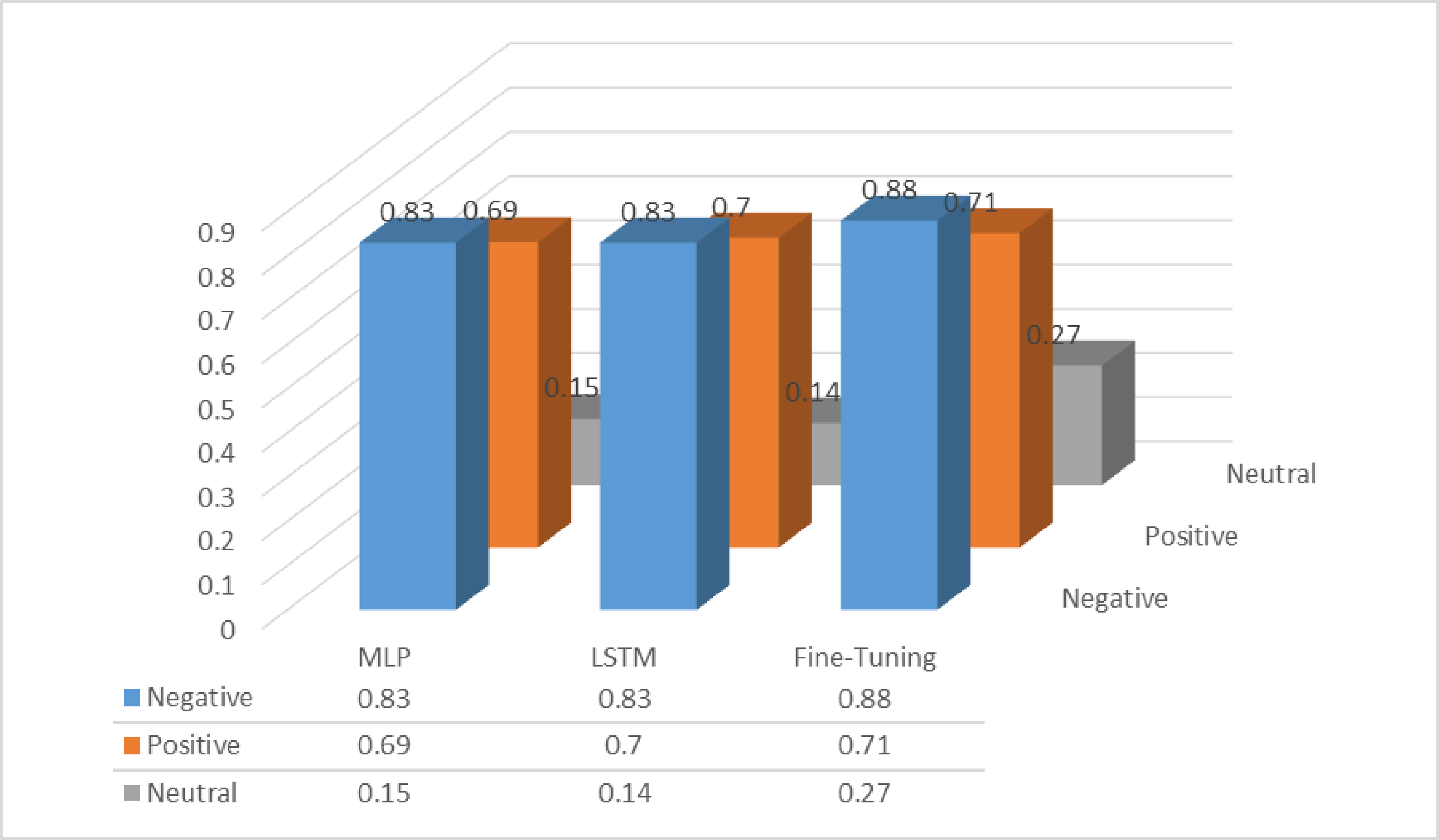

Figure 14: F1 measure of various classifiers for BERT Model4 for each class

### 4.1.5 Summary of BERT Models

All the results mentioned above are shown in Figure 15 together here which better shows that fine-tuning achieved very good results in all cases.

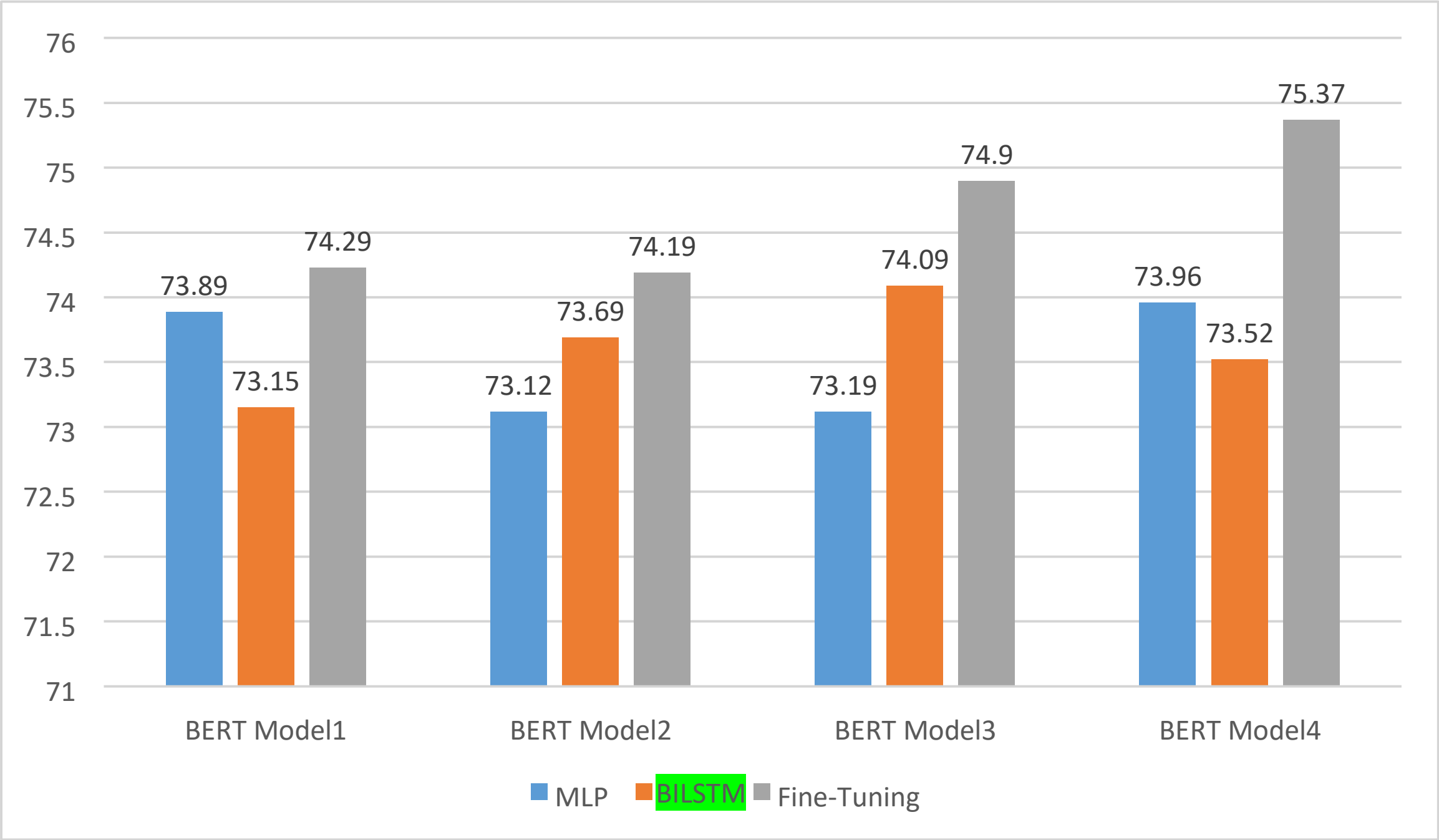

Figure 15: Accuracy results for all BERT Models

## 4.2 Comparison with Previous Works

For comparison, we have used our previous work (Muhealddin and Veisi, 2022) as the baseline. In that work, the Word2Vec representation was used which was trained in two ways and the best case was the vector size of 100 and the window size was 5 with the BiLSTM classifier whose accuracy was %71.35. For both word embedding techniques, we used the same classifier (i.e., the BILSTM) to find the best word embedding method. For BERT word embedding with the BILSTM classifier, the best performance was %74.09. As shown in Figure 16, the BERT model of the same classifier shows better performance than the Word2Vec, if we take the best case of BERT, the accuracy is 75.37%, and for Word2Vec, the best-case accuracy is 71.35% as shown in Figure 17.

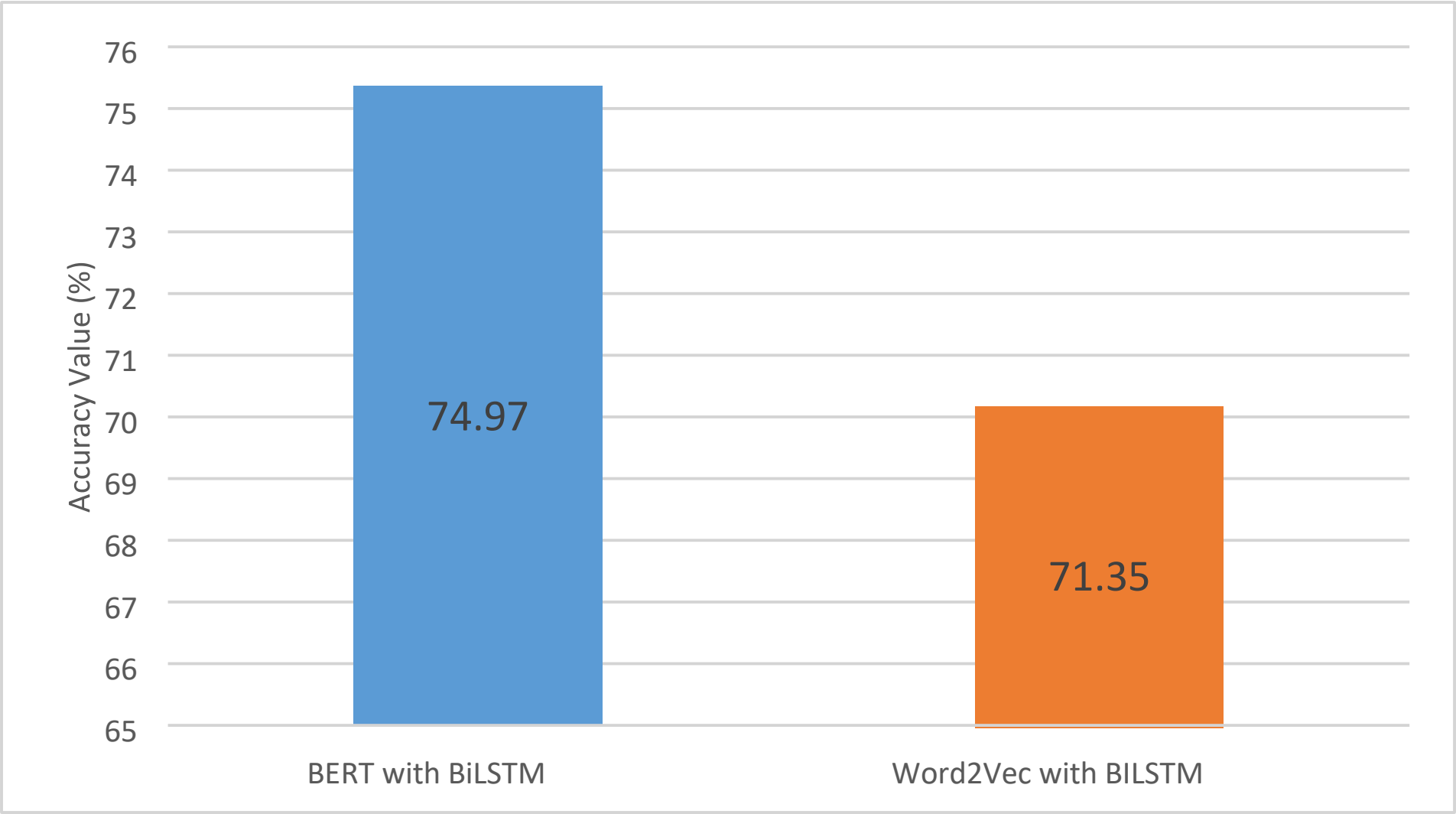

Figure 16: Comparison between Word2Vec and BERT by using BILSTM Classifier

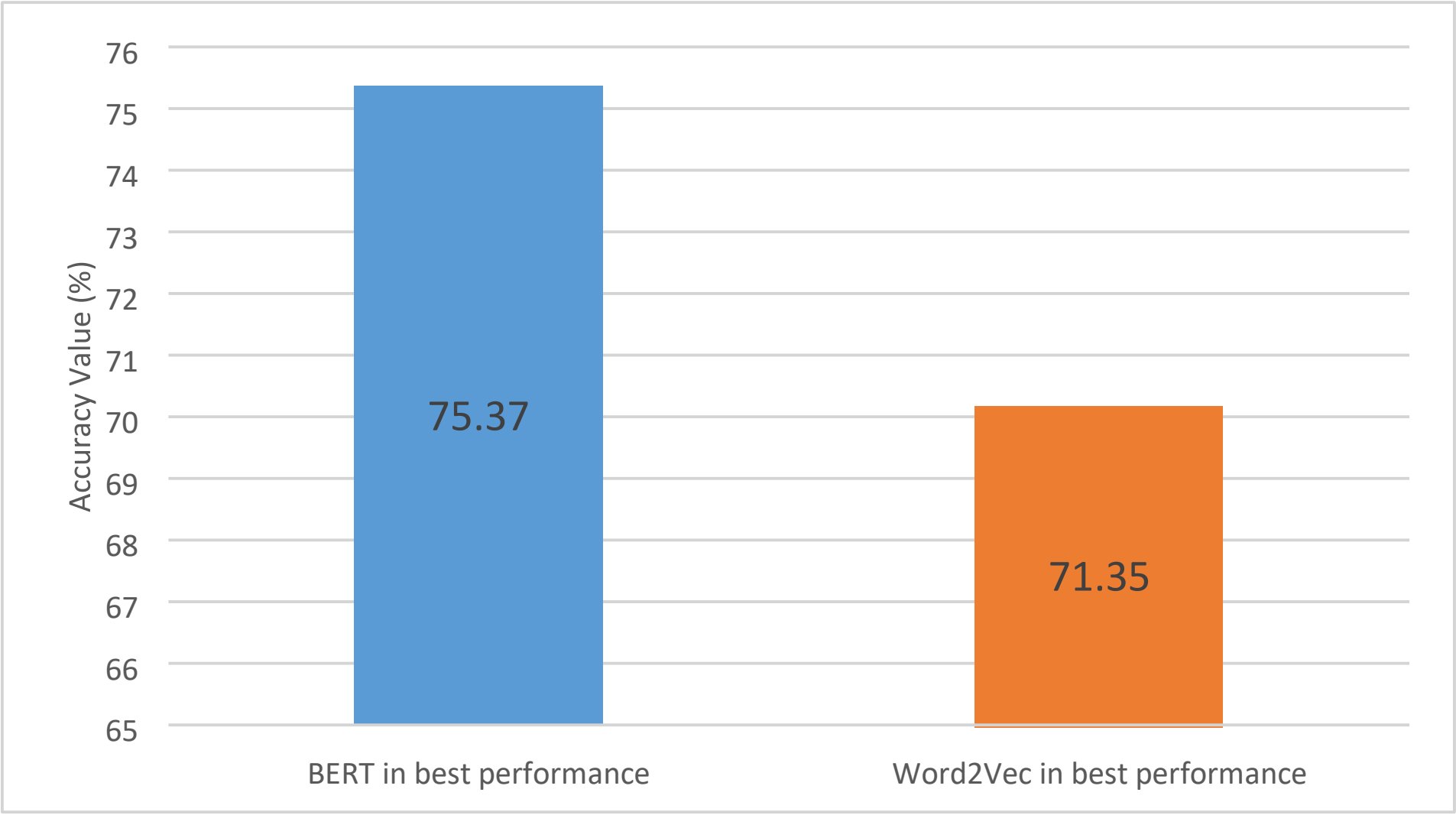

Figure 17: Comparison between Word2Vec and BERT in the best performance

In addition, we evaluated both the BERT and Word2Vec models on the KurdiSent dataset shown in Figure 18, an annotated dataset containing more than 12,000 instances used to analyze sentiments in Kurdish (positive and negative categories). The BERT model achieved an accuracy of 87.05%, while the Word2Vec model achieved 82.01%. These results clearly show that the BERT-based approach significantly outperforms Word2Vec, which further highlights the effectiveness of incorporating the energy transformation principle for sentiment analysis in Kurdish.

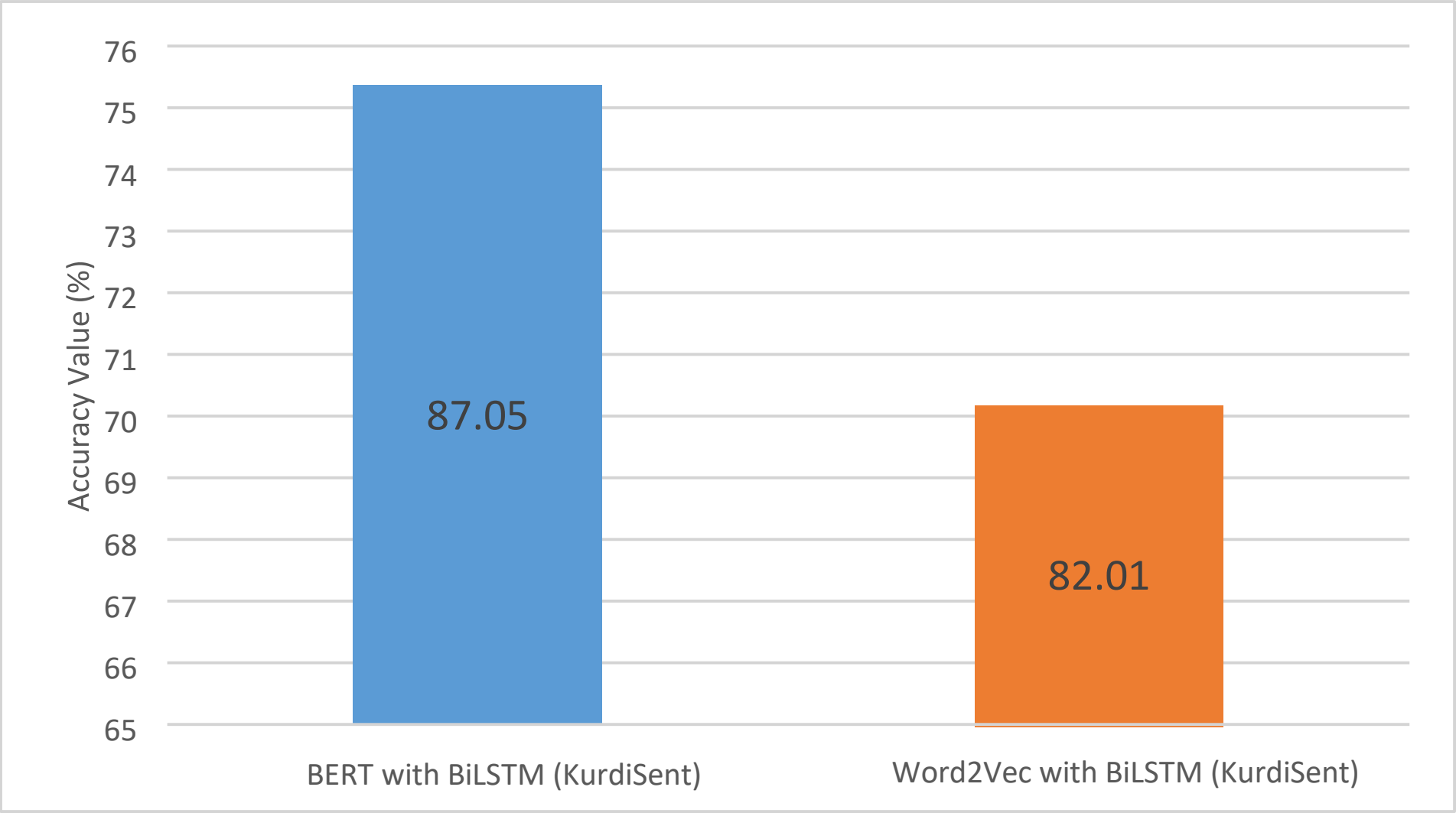

Figure 18 Comparison between Word2Vec and BERT in KurdiSent corpus

## 4.3 2-Classes Sentiment Analysis Model

As mentioned before, the labeled dataset is imbalanced as shown in Figure 1, which greatly reduces the accuracy of the models. The dataset has a much less neutral class than negative class, which means that the trained deep model will have a weak learning ability for the neutral class and have the wrong prediction for this class. Therefore, it was considered to remove the neutral samples and conduct a test of the best language model, which had an accuracy of 75.37% for three classes. As it is shown in Table 12 the accuracy of the BERT model in this case is 86.31%. This is because the data is less imbalanced than before, however, the imbalance rate is still high as shown in Figure 1. Certainly, if trained entirely on balanced data (using under-sampling method), the results will be much higher. The micro average for BERT 2-classes model is 87%.

Table 12: Comparison between 3-Classes and 2-Classes for BERT-BILSTM

| Model | Accuracy |
|---|---|
| BERT 3-Classes in best Case | **75.37%** |
| BERT 2-Classes in best Case | **86.31%** |

# 5. Summary and Conclusions

In this study we presented an advancement in sentiment analysis for the Kurdish language by integrating BERT language model with NLP techniques. Through meticulous methodology, encompassing the collection and normalization of a substantial Kurdish text corpus, the employment of BERT's sophisticated word embedding capabilities, and the application of advanced classifiers, this research sets a new benchmark in understanding and analyzing sentiments within the Kurdish linguistic context. The study's findings underscore the efficacy of BERT over traditional word embedding techniques like Word2Vec, demonstrating a notable improvement in accuracy and the ability to capture nuanced semantic relationships. This is particularly important for a low-resource language like Kurdish, where data scarcity and linguistic complexity pose significant challenges. The adaptation of BERT, complemented by BILSTM, MLP, and fine-tuning approaches, has not only addressed these challenges but also highlighted the potential for deep learning models to enhance sentiment analysis in languages with limited computational resources. Furthermore, the comparison between the performance of BERT and Word2Vec, particularly using BILSTM classifiers, illustrates the transformative impact of leveraging pre-trained models and fine-tuning techniques in NLP tasks. The improved accuracy and F1 measures across various models affirm the superiority of BERT in handling complex linguistic phenomena and sentiment analysis. This research contributes to the field of NLP by paving the way for future studies focused on low-resource languages, encouraging the exploration of LLMs in languages beyond Kurdish. Additionally, the development of a sentiment analysis model that effectively utilizes BERT for Kurdish represents a crucial step towards bridging the gap between advanced NLP techniques and the needs of low-resource linguistic communities.

Following the development of 12 distinct BERT models tailored for sentiment analysis, particularly emphasizing the Kurdish language. As a result, is the pinnacle accuracy rate of 75.37% attained by the Fine-Tuning classifier. This achievement exemplifies BERT's exceptional proficiency in sentiment analysis, surpassing the Word2Vec method. It accentuates the advantage of utilizing sophisticated language models to efficiently process and analyze sentiment in textual data.

As a result., the limitation of the study is the asymmetric data, especially in the neutral sentiment category, which affects the performance of the model. Future work should focus on expanding and balancing the datasets by combining more diverse sources and using data augmentation techniques. In addition, we should use advanced models such as RoBERTa and LLaMA for Kurdish in the future, but they require large data sets, which can improve the results. Expanding the use of KuBERT to other NLP tasks, such as text summarization and machine translation, is also recommended for future development.